%% file: main.tex
\documentclass{article}  %
\usepackage{iclr2026_conference,times}

\input{math_commands.tex}

\usepackage{capt-of, eucal}
\usepackage{xspace}
\usepackage{xcolor}
\usepackage{enumitem}
\usepackage{amsmath,amsthm,amssymb,amsfonts,dsfont,pifont,bm,bbm,mathrsfs,mathtools,nicefrac,extarrows,relsize}
\usepackage{algorithm,algpseudocode,listings}
\usepackage{booktabs,multirow,adjustbox,diagbox,threeparttable,tabularray,setspace}
\usepackage{wrapfig}
\usepackage{graphicx}
\usepackage{subcaption}
\usepackage{tabularx}
\usepackage{array}
\usepackage{makecell}
\usepackage{longtable} 
\usepackage{paracol}
\usepackage{float}
\usepackage{fontawesome5}
\usepackage[tableposition=top]{caption}
\usepackage{colortbl}
\usepackage{color}
\usepackage{multicol}
\usepackage{lipsum}

\definecolor{citeblue}{rgb}{0.21,0.49,0.74}
\usepackage[pagebackref=false,breaklinks,colorlinks,citecolor=citeblue,bookmarks=false]{hyperref}
\usepackage{wrapfig}
\usepackage[capitalize]{cleveref}  %

\crefname{section}{Sec.}{Secs.}
\Crefname{section}{Section}{Sections}
\crefname{appendix}{Appendix}{Appendices}
\Crefname{appendix}{Appendix}{Appendices}
\crefname{table}{Table}{Tables}
\Crefname{table}{Table}{Tables}
\crefname{figure}{Fig.}{Figs.}
\Crefname{figure}{Figure}{Figures}
\crefname{equation}{Eq.}{Eqs.}
\Crefname{equation}{Equation}{Equations}
\crefname{theorem}{Thm.}{Thms.}
\Crefname{theorem}{Theorem}{Theorems}
\crefname{lemma}{Lem.}{Lems.}
\Crefname{lemma}{Lemma}{Lemmas}
\crefname{remark}{Rem.}{Rems.}
\Crefname{remark}{Remark}{Remarks}
\crefname{corollary}{Cor.}{Cors.}
\Crefname{corollary}{Corollary}{Corollaries}
\crefname{algorithm}{Alg.}{Algs.}
\Crefname{algorithm}{Algorithm}{Algorithms}
\definecolor{cellred}{RGB}{213, 123, 101}
\definecolor{cellgreen}{RGB}{0, 205, 0}
\definecolor{cellblue}{RGB}{54, 125, 189}
\definecolor{codegreen}{rgb}{0,0.6,0}
\definecolor{codegray}{rgb}{0.5,0.5,0.5}
\definecolor{codepurple}{rgb}{0.58,0,0.82}
\definecolor{backcolour}{rgb}{1.0,1.0,1.0}
\lstdefinestyle{mystyle}{
    backgroundcolor=\color{backcolour},
    commentstyle=\color{codegreen},
    keywordstyle=\color{magenta},
    numberstyle=\tiny\color{codegray},
    stringstyle=\color{codepurple},
    basicstyle=\ttfamily\scriptsize,
    breakatwhitespace=false,
    breaklines=true,
    captionpos=b,
    keepspaces=true,
    numbers=left,
    numbersep=5pt,
    showspaces=false,
    showstringspaces=false,
    showtabs=false,
    tabsize=2
}
\usepackage[most,skins,theorems]{tcolorbox}
\tcbset{
  aibox/.style={
    width=\linewidth,
    top=8pt,
    bottom=4pt,
    colback=blue!6!white,
    colframe=black,
    colbacktitle=black,
    enhanced,
    center,
    attach boxed title to top left={yshift=-0.1in,xshift=0.15in},
    boxed title style={boxrule=0pt,colframe=white,},
  }
}
\newtcolorbox{AIbox}[2][]{aibox,title=#2,#1}
\usepackage{amsmath} %
\usepackage{amssymb} %
\usepackage{multirow}

\usepackage{rotating}  %

\newcolumntype{C}[1]{>{\centering\arraybackslash}p{#1}}
\newcolumntype{L}[1]{>{\arraybackslash}p{#1}}

\definecolor{demphcolor}{gray}{.2}

\definecolor{demphcolorinline}{gray}{.3}

\definecolor{demphcolor1}{gray}{.6}

\newcommand{\tocite}[1]{{\color{red} [TO CITE]}}

\newcommand{\methodname}{SpatialLadder}
\newcommand{\datasetname}{SpatialLadder-26$k$}

\definecolor{LightCyan}{rgb}{0.94, 1.0, 1.0}
\definecolor{ForestGreen}{rgb}{0.13, 0.55, 0.13}
\definecolor{LightBlue}{rgb}{0.678, 0.847, 0.902}
\definecolor{DarkRed}{rgb}{0.7, 0.1, 0.1}

 \title{\raisebox{-0.1ex}{\adjustbox{height=2.0em, valign=m}{\includegraphics[height=2.0em]{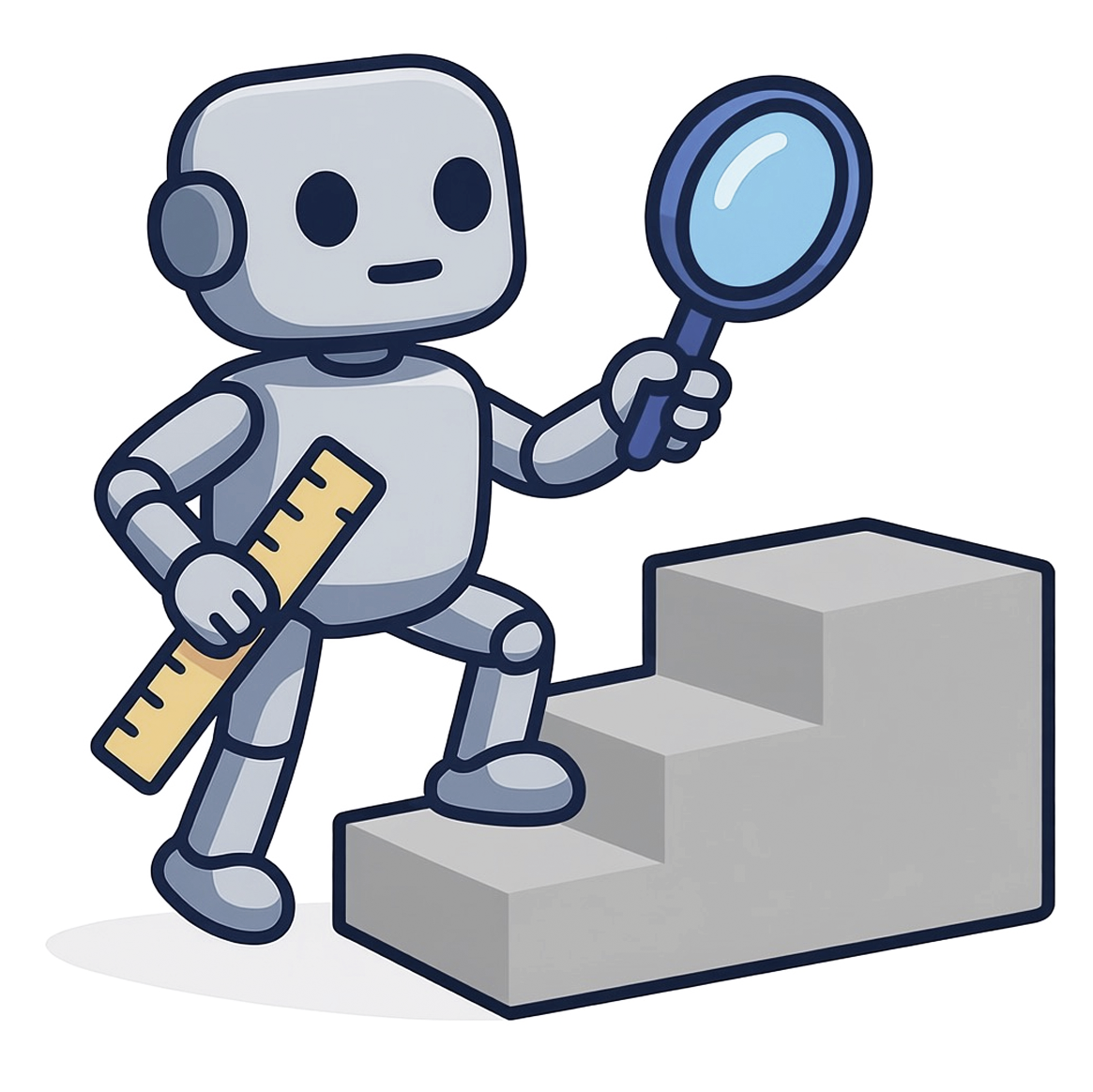}}}
 SpatialLadder: Progressive Training for Spatial Reasoning in Vision-Language Models\vspace{-10pt}}

\author{%
  \textbf{Hongxing Li}$^{1,*}$,
  ~~
  \textbf{Dingming Li}$^{1,*}$
  ~~
  \textbf{Zixuan Wang}$^{1}$
  ~~
  \textbf{Yuchen Yan}$^{1}$
  ~~
  \textbf{Hang Wu}$^{1}$ \\
  \textbf{Wenqi Zhang}$^{1}$
  ~~
  \textbf{Yongliang Shen}$^{1,\dagger}$,
  ~~
  \textbf{Weiming Lu}$^{1}$,
  ~~
  \textbf{Jun Xiao}$^{1}$
  ~~
  \textbf{Yueting Zhuang}$^{1}$ \\
  $^1$Zhejiang University\\
  \texttt{\{hongxing.li, syl\}@zju.edu.cn} \\
  \begin{tabular}{@{}ll@{}}
  \vspace{-5pt} \\
    \faGithub\ GitHub: & \href{https://github.com/zju-real/SpatialLadder}{\texttt{\textcolor{cyan}{https://github.com/ZJU-REAL/SpatialLadder}}} \\
    \faGlobe\ Project: & \href{https://zju-real.github.io/SpatialLadder}{\texttt{\textcolor{cyan}{https://zju-real.github.io/SpatialLadder}}}
  \end{tabular}
}

\iclrfinalcopy %

\begin{document}

\maketitle

\vspace{-10pt}
\input{sections/0.abs}

\vspace{-10pt}
\input{sections/1.intro}

\vspace{-10pt}
\input{sections/3.related_work}

\vspace{-10pt}
\input{sections/4.method}

\input{sections/5.experiments}

\input{sections/6.conclusion}

\input{sections/8.statement}

\input{sections/9.bib}

\input{sections/99.appendix}

\end{document}

%% file: math_commands.tex
\usepackage{amsmath,amsfonts,bm}

\def\eqref#1{equation~\ref{#1}}

\def\1{\bm{1}}

\DeclareMathAlphabet{\mathsfit}{\encodingdefault}{\sfdefault}{m}{sl}
\SetMathAlphabet{\mathsfit}{bold}{\encodingdefault}{\sfdefault}{bx}{n}

%% file: sections/0.abs.tex
\vspace{-5pt}
\begin{abstract}

Spatial reasoning remains a fundamental challenge for Vision-Language Models (VLMs), with current approaches struggling to achieve robust performance despite recent advances. We identify that this limitation stems from a critical gap: existing methods attempt to learn spatial reasoning directly without establishing the hierarchical foundations of perception and understanding. To address this challenge, we present a comprehensive methodology for building spatial intelligence progressively. We introduce \datasetname, a multimodal dataset containing 26,610 samples spanning object localization, single-image, multi-view, and video spatial reasoning tasks, constructed through a standardized pipeline that ensures systematic coverage across modalities. Building on this dataset, we design a three-stage progressive training framework that (1) establishes spatial perception through object localization, (2) develops spatial understanding through multi-dimensional spatial tasks, and (3) strengthens complex reasoning via reinforcement learning with verifiable rewards. This approach yields \methodname, a 3B-parameter model that achieves state-of-the-art performance on spatial reasoning benchmarks, with 23.4\% average improvement over the base model, surpassing GPT-4o by 20.8\% and Gemini-2.0-Flash by 10.1\%. Notably, \methodname{} maintains strong generalization with 7.2\% improvement on out-of-domain benchmarks, demonstrating that progressive training from perception to reasoning is essential for robust spatial intelligence.

\end{abstract}

%% file: sections/1.intro.tex
\section{Introduction}

VLMs have achieved remarkable success in fundamental visual tasks~\citep{huang2025vision-r1, yu2025perception}, yet a critical capability remains elusive: spatial reasoning. While humans effortlessly understand spatial relationships in visual scenes, current VLMs struggle with even basic spatial queries~\citep{yang2025thinking, tong2024cambrian, wu2025spatialscore}. This limitation severely constrains their deployment in applications requiring spatial intelligence, from robotics navigation~\citep{zitkovich2023rt} to autonomous driving~\citep{tian2024drivevlm} and virtual reality systems~\citep{chandrasegaran2024hourvideo}.

The root cause of this spatial reasoning deficit lies in a fundamental gap between perception and reasoning in current VLM architectures~\citep{chen2025perceptionreasoningtwostagereinforcement, li2025self-rewarding}. We hypothesize that existing approaches fail because they treat spatial reasoning as a monolithic capability, attempting to learn it directly from question-answer pairs without establishing the necessary hierarchical structure~\citep{ouyang2025spacer, wu2025spatial}.
To validate this hypothesis, we conducted controlled experiments with 200 spatial orientation tasks, progressively adding perceptual hints to isolate the bottleneck (detailed in Appendix \ref{app:preliminary}). As shown in Figure \ref{fig:preliminary}, providing location hints (bounding boxes) improves accuracy by 5.0\%, and additional directional cues yield another 4.5\% gain, achieving 9.5\% total improvement.
This demonstrates that models possess latent reasoning capabilities but lack the perceptual grounding to activate them effectively. The primary bottleneck lies not in reasoning capacity but in the integration between perception and reasoning.

Current approaches to enhancing spatial reasoning in VLMs suffer from two fundamental limitations. First, existing datasets are fragmented and narrow in scope, focusing on either 2D images or 3D scenes in isolation~\citep{liao2025improved, ouyang2025spacer, kamath2023s}, while lacking systematic coverage across modalities and standardized annotation pipelines, resulting in incomplete training signals for comprehensive spatial understanding. Second, recent methods attempt to directly optimize reasoning outputs through reinforcement learning~\citep{liao2025improved, ouyang2025spacer, wu2025reinforcing} or auxiliary 3D representations~\citep{wu2025spatial, hong20233d, zhu2024llava-3d, zheng2025video-3d}, without establishing the hierarchical structure required for spatial intelligence: they bypass the critical progression from perceiving objects to understanding spatial relationships to performing logical inference, producing models that memorize patterns rather than develop genuine spatial understanding, leading to poor generalization on novel spatial configurations.

We address these challenges through a systematic approach based on the hierarchical nature of spatial intelligence. Our key insight is that robust spatial reasoning must be built progressively: establishing perceptual foundations through object localization, developing spatial understanding through multi-dimensional spatial analysis, and ultimately achieving complex reasoning.

\begin{figure}
    \centering
    \includegraphics[width=\linewidth]{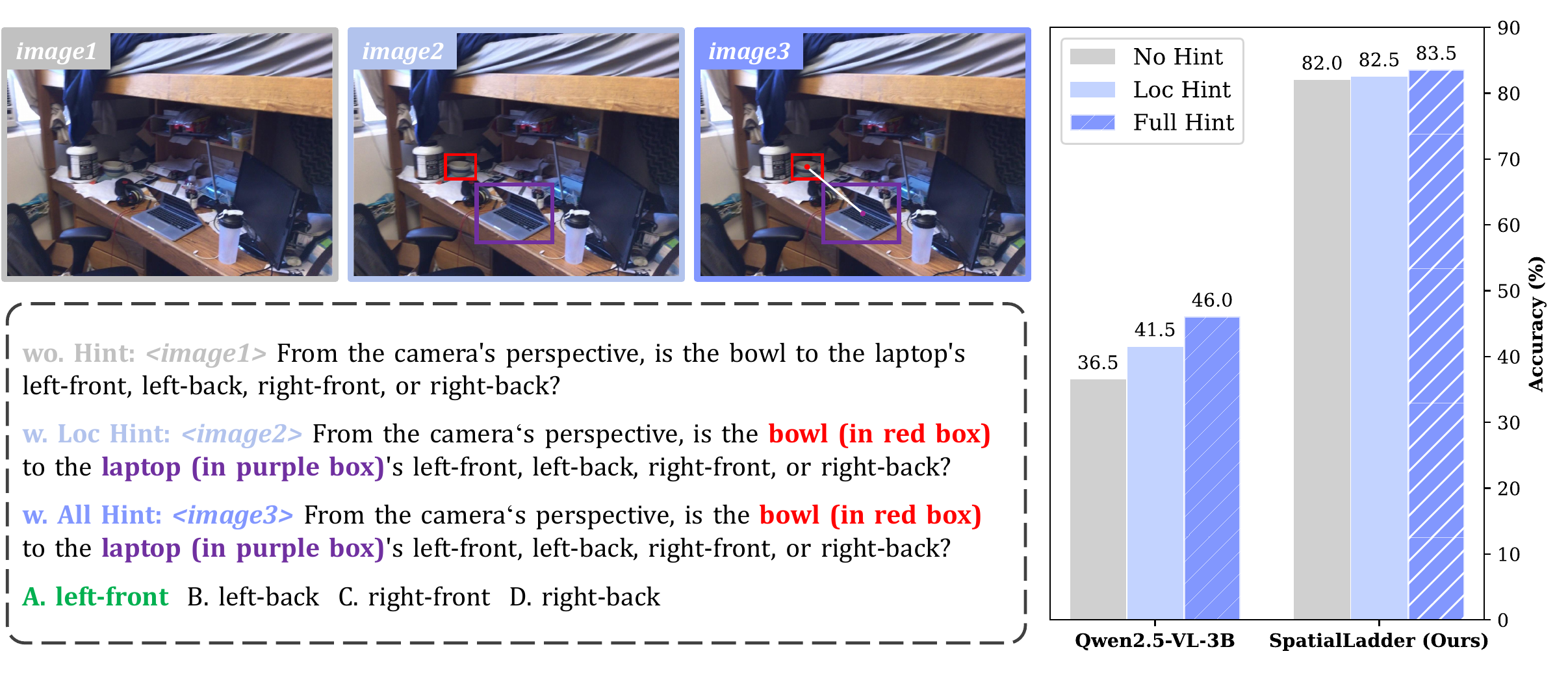}
    \caption{\textbf{Perception-reasoning gap in spatial reasoning.} Left: Three experimental conditions with increasing perceptual hints: no hints, location hints (bounding boxes), and full hints (boxes plus directional cues). Right: While Qwen2.5-VL-3B shows progressive improvement with increasing hints, our trained model achieves superior performance with negligible reliance on external prompts.}
    \label{fig:preliminary}
    \vspace{-14pt}
\end{figure}

To implement this vision, we introduce \datasetname, a comprehensive multimodal dataset containing 26,610 samples across four complementary task categories: object localization (5,929 samples), single-image spatial reasoning (5,929 samples), multi-view spatial reasoning (5,752 samples), and video spatial reasoning (9,000 samples). Unlike existing datasets, \datasetname{} systematically covers the full spectrum from basic perception to complex reasoning. We develop a standardized pipeline leveraging 3D scene reconstructions from ScanNet to ensure consistent, high-quality annotations across all modalities.

Building on this dataset, we design a three-stage progressive training framework. Stage 1 establishes spatial perception through object localization tasks, teaching models to accurately identify and locate objects in scenes. Stage 2 develops spatial understanding through multi-dimensional tasks including size estimation, distance judgment, and orientation analysis across seven distinct spatial dimensions. Stage 3 employs Group Relative Policy Optimization (GRPO)~\citep{shao2024deepseekmath} with task-specific verifiable reward functions to strengthen complex reasoning capabilities, enabling models to form coherent chains of spatial thought.

Through this progressive approach, we develop \methodname, a 3B-parameter model that establishes new benchmarks in spatial reasoning performance. Extensive experiments demonstrate significant improvements: on VSI-Bench~\citep{yang2025thinking}, \methodname{} achieves 45.7\% accuracy. On our proposed SPBench-SI and SPBench-MV benchmarks, it attains 70.2\% and 70.9\% accuracy respectively. Across all benchmarks, \methodname{} achieves an overall performance of 62.3\%, surpassing the base model by 23.4\% and outperforming GPT-4o by 20.8\% and Gemini-2.0-Flash by 10.1\%. Crucially, \methodname{} maintains strong generalization with 7.2\% average improvement on out-of-domain benchmarks including CV-Bench~\citep{tong2024cambrian}, SPAR~\citep{zhang2025flatland}, and ViewSpatial-Bench~\citep{li2025viewspatial}, demonstrating the robustness of our progressive training approach.

Our contributions are threefold:
\begin{itemize}
\item We introduce \datasetname, a comprehensive multimodal dataset with 26,610 samples spanning object localization and spatial reasoning across single-image, multi-view, and video modalities, constructed through a standardized pipeline ensuring systematic coverage and high-quality annotations.
\item We design a three-stage progressive training framework that systematically builds spatial reasoning capabilities by establishing perceptual foundations, developing spatial understanding, and strengthening complex reasoning through reinforcement learning with verifiable rewards.
\item We demonstrate that our approach yields significant performance improvements, with \methodname{} achieving state-of-the-art results on multiple benchmarks while maintaining strong generalization to out-of-domain tasks, validating the effectiveness of progressive spatial learning.
\end{itemize}

%% file: sections/3.related_work.tex
\section{Related Works}

\subsection{Visual Spatial Reasoning}
As a key capability of VLMs, visual spatial reasoning is more complex than general visual tasks and remains challenging~\citep{yang2025thinking, wu2025spatialscore}. Despite notable advances in basic visual tasks~\citep{li2024llava-one, li2024llava-nxt}, extensive benchmark~\citep{yang2025thinking, wu2025spatialscore, li2025sti} evaluations demonstrate that they still face serious bottlenecks in spatial reasoning. Recent studies have attempted to explore multiple remedies, such as R1-Zero-VSI~\citep{liao2025improved} and SpaceR~\citep{ouyang2025spacer}, which utilize reinforcement learning to enhance models' spatial reasoning capabilities; Spatial-MLLM~\citep{wu2025spatial}, which introduces 3D representations~\citep{wang2025vggt} as bridging knowledge; and Coarse Correspondences~\citep{liu2025coarse}, which improves models' spatiotemporal modeling capabilities through cross-frame object tracking. However, there remains a general lack of comprehensive, diverse, high-quality datasets, as well as effective training frameworks that advance from basic to complex concepts, to systematically enhance the capabilities of VLMs in spatial reasoning tasks.

\subsection{Reinforcement Learning in VLMs}
Recent studies have extended Reinforcement Learning (RL) techniques from LLMs to VLMs, leading to notable progress in visual reasoning~\citep{liu2025visual-rft, shen2025vlm-r1}. Representative works such as Vision-R1~\citep{huang2025vision-r1}, MM-Eureka~\citep{meng2025mm}, and R1-OneVision~\citep{yang2025r1-onevision} have demonstrated that transferring RL methods to VLMs can significantly enhance the visual mathematical reasoning capabilities. In the video domain, Video-R1~\citep{feng2025video} and VideoChat-R1~\citep{li2025videochat} applied RL to improve temporal understanding and video localization performance. Beyond text-oriented reasoning, methods like GRIT~\citep{fan2025grit} and Pixel-Reasoner~\citep{su2025pixel} leveraged RL to stimulate “thinking with images”~\citep{su2025thinking_with_image}, enabling models to perform structured and interpretable multimodal reasoning. Despite these advancements, research specifically targeting visual spatial reasoning remains limited. To address this gap, we propose a training paradigm that systematically enhances VLMs’ capabilities in spatial reasoning tasks.

%% file: sections/4.method.tex
\section{Methods}
We present a comprehensive framework that systematically builds spatial reasoning in VLMs via progressive training. Our approach consists of two core components: (1) \datasetname, a multimodal dataset systematically spanning spatial tasks from basic perception to complex reasoning, and (2) a three-stage training framework reflecting the hierarchical nature of spatial intelligence.

\subsection{Dataset Construction}
\begin{figure}[t]
	\centering
	\includegraphics[width=1\linewidth]{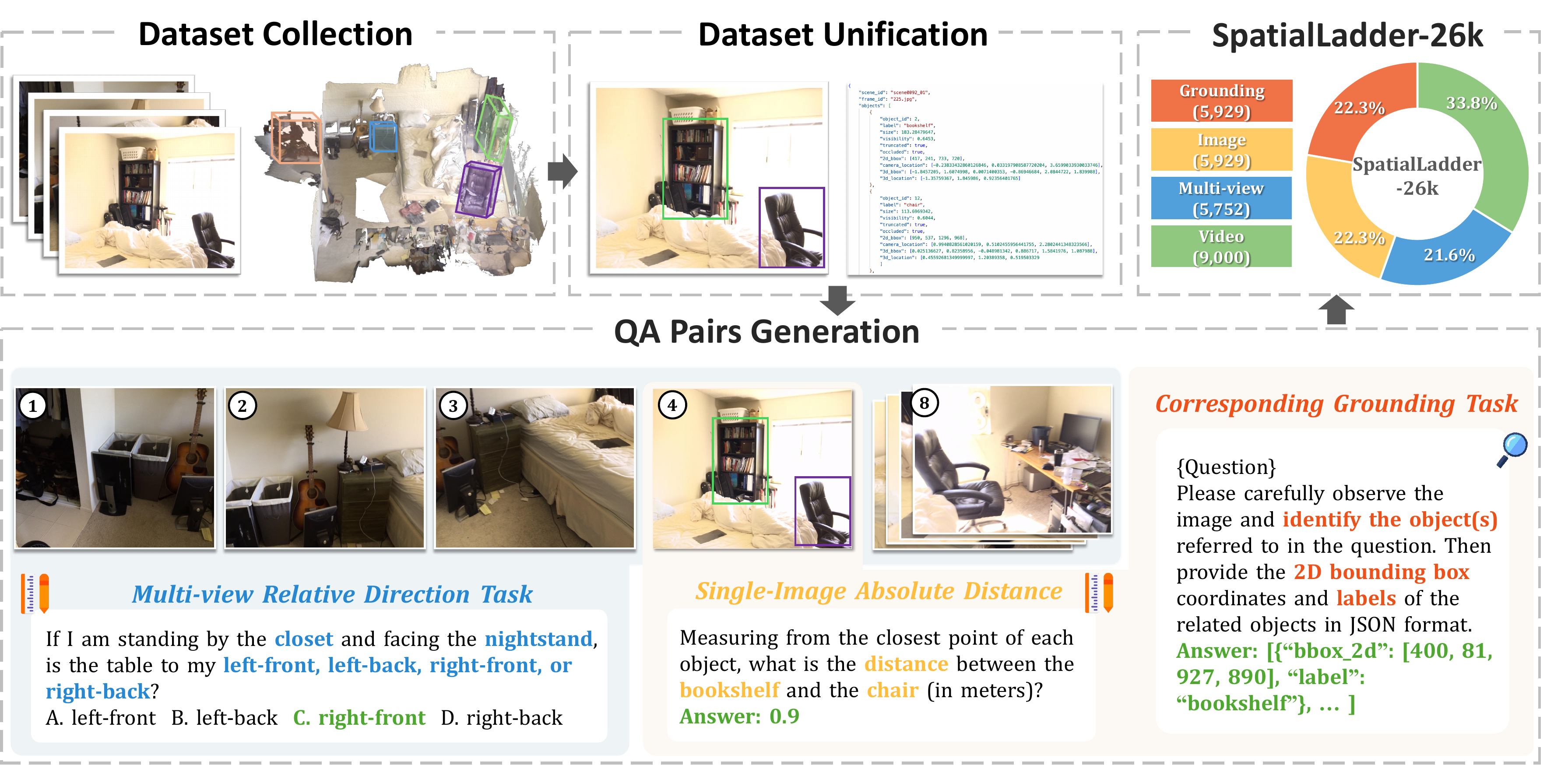}
	\caption{\textbf{Overview of \datasetname{} dataset construction pipeline} from raw data collection to question–answer pairs generation, with representative tasks including multi-view relative direction, single-image absolute distance, and corresponding grounding tasks.}
	\label{fig:dataset}
    \vspace{-10pt}
\end{figure}

Effective spatial reasoning requires diverse, high-quality training data spanning from basic perception to complex reasoning. We introduce \datasetname, comprising 26,610 samples across four complementary task categories that form a complete spatial learning curriculum. Figure~\ref{fig:dataset} illustrates our construction pipeline and dataset composition.

\paragraph{Task Design and Hierarchy.} 
Our strategically designed dataset comprises four task categories: object localization (5,929 samples), single-image spatial reasoning (5,929 samples), multi-view spatial reasoning (5,752 samples), and video spatial reasoning (9,000 samples). Object localization establishes perceptual foundations via precise bounding box predictions for spatially-referenced objects. Spatial reasoning tasks span three modalities and seven dimensions: relative direction, relative distance, absolute distance, object size, counting, room size, and appearance order. Single-image tasks provide the entry point for static scene reasoning. Multi-view tasks require cross-perspective integration, synthesizing eight distinct viewpoints of identical environments. Video tasks incorporate temporal dynamics through 1–4 minute sequences at 24 fps, demanding coherent spatiotemporal understanding. This hierarchical progression ensures systematic capability development from foundational perception to complex spatiotemporal reasoning.

\paragraph{Construction Pipeline.}
Figure~\ref{fig:dataset} details our standardized three-stage pipeline to ensure systematic data generation across all modalities. In the first stage, we collect ScanNet’s~\citep{dai2017scannet} comprehensive 3D scene reconstructions for object localization, single-image spatial reasoning, and multi-view spatial reasoning, and carefully sample 9,000 videos from SR-91k~\citep{ouyang2025spacer} to support video spatial reasoning. In the second stage, we perform 3D-to-2D transformations and dataset unification, obtaining rich information including 3D bounding boxes, 2D bounding boxes, 3D absolute locations, 2D locations relative to the camera, visibility ratios and object sizes. In the third stage, we generate diverse question–answer pairs using templates adapted from VSI-Bench~\citep{yang2025thinking} to construct tasks across different spatial reasoning scenarios. Further details on dataset construction (e.g. quality assurance, QA templates) are provided in~\ref{app:dataset_construction}.

\subsection{Three-stage Progressive Training Framework}

\begin{figure}[t!]
	\centering
	\includegraphics[width=1\linewidth, trim={0 190pt 0 0}, clip]{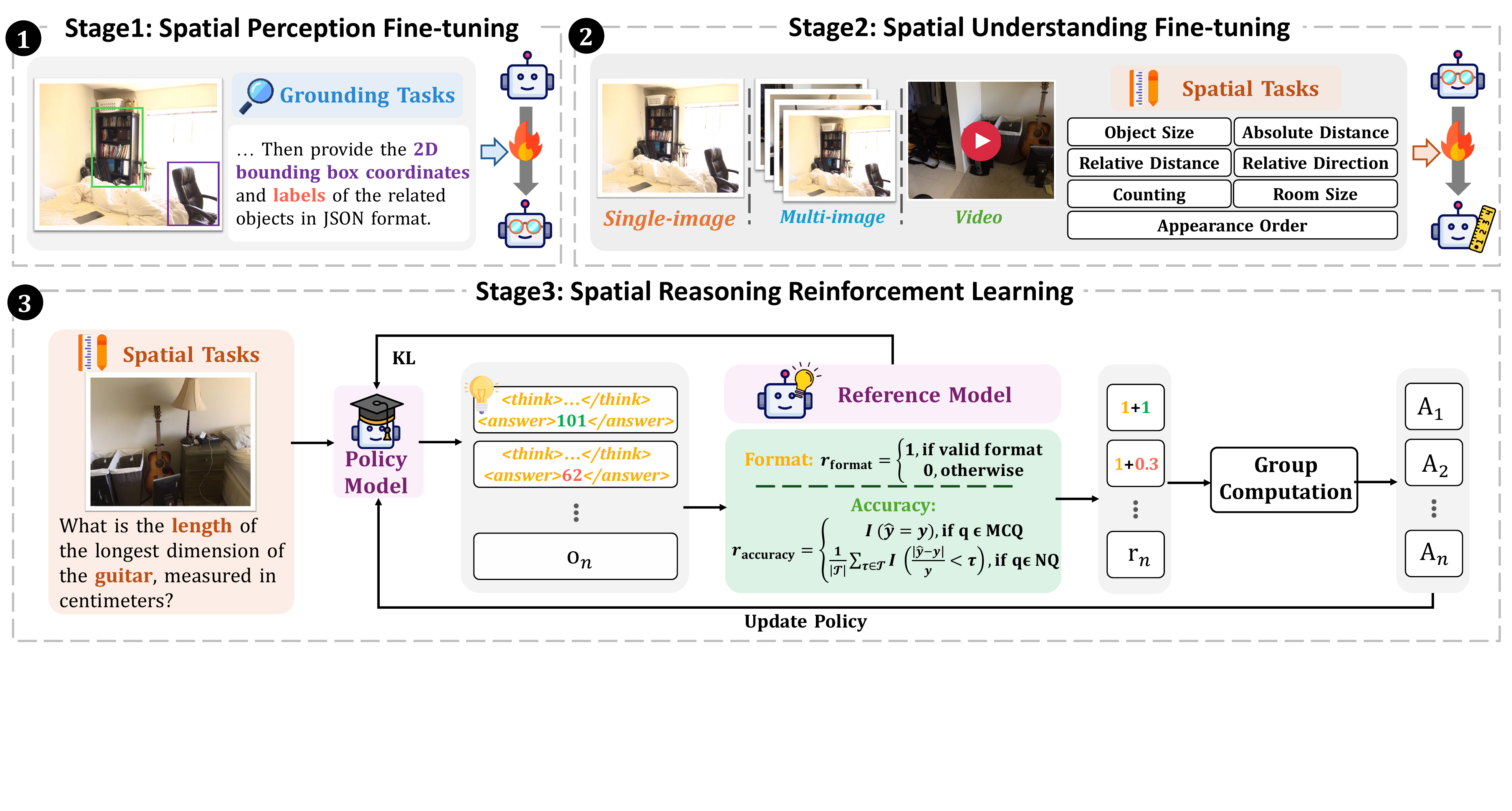}
	\caption{\textbf{Three-stage progressive training framework of \methodname.} Stage 1 establishes perceptual grounding through object localization, Stage 2 develops spatial understanding across seven dimensions using multimodal tasks, and Stage 3 employs GRPO reinforcement learning with chain-of-thought generation to strengthen reasoning capabilities.}
	\label{fig:framework}
    \vspace{-10pt}
\end{figure}

Building upon \datasetname, we design a training framework that systematically constructs spatial intelligence through three progressive stages, as illustrated in Figure~\ref{fig:framework}, each addressing a specific level of the spatial reasoning hierarchy. The framework embodies the principle that robust spatial reasoning emerges from the integration of perception, understanding, and reasoning, with each stage building upon foundations established in previous stages.

\paragraph{Stage 1: Perceptual Grounding through Localization.}
The first stage establishes foundational spatial perception via object localization on 6k \datasetname\ samples. The model learns to link visual inputs with spatial queries, producing JSON outputs containing object identities and 2D bounding boxes. This stage grounds abstract spatial concepts in concrete visual evidence. Through supervised fine-tuning, the model develops three core capabilities: distinguishing spatially relevant objects from background elements, robust detection tailored to spatial reasoning contexts, and mappings between linguistic descriptions and visual regions. The training emphasizes localization precision, as accurate object detection underpins all subsequent spatial reasoning. By focusing exclusively on perceptual tasks, we ensure strong visual grounding before advancing to complex reasoning.

\paragraph{Stage 2: Spatial Understanding through Multi-dimensional Tasks.} 
The second stage broadens spatial comprehension by introducing comprehensive reasoning tasks that include size estimation, distance judgment, and orientation analysis across seven distinct spatial dimensions: relative direction, relative distance, absolute distance, object size, counting, room size, and appearance order. Training spans three modalities with distinct contributions: single-image tasks establish fundamental spatial relationships, multi-view tasks demand cross-perspective integration and implicit 3D understanding, while video tasks add temporal dynamics and motion tracking capabilities. This multimodal approach creates robust spatial representations that generalize across visual contexts. The supervised fine-tuning requires flexible adaptation between multiple-choice questions testing discrete concepts and numerical questions demanding precise measurements, developing comprehensive spatial understanding that transcends individual task types.

\paragraph{Stage 3: Spatial Reasoning through Reinforcement Learning.} 
The final stage transforms spatial understanding into explicit reasoning capabilities through reinforcement learning with chain-of-thought~\citep{wei2022cot} generation. We implement a carefully designed reward structure that evaluates both reasoning quality and answer correctness:
\begin{equation}
\mathcal{R}(o, y) = r_{\text{format}}(o) + r_{\text{accuracy}}(o, y)
\end{equation}
Format rewards ensure structured reasoning by checking for proper \texttt{<think>} and \texttt{<answer>} tag usage, encouraging the model to explicitly articulate its reasoning process. Accuracy rewards are task-specific: binary for multiple-choice questions and graduated for numerical answers based on relative error thresholds. This dual reward structure prevents the model from generating plausible-sounding but incorrect reasoning chains.

We employ GRPO for stable policy optimization. For each question $q$, the model samples a series of candidate answers $\{o_1,o_2,...,o_G\}$ from the policy model $\pi_\mathrm{old}$ and optimizes the policy by maximizing the following objective function:
\begin{equation}
\small
\mathcal{J}_{\text{GRPO}}(\theta) = \mathbb{E}_{q,o_i} \left[ \frac{1}{G} \sum_{i=1}^{G} \min \left( \frac{\pi_\theta(o_i | q)}{\pi_{\theta_{\text{old}}}(o_i | q)} A_i, \text{clip}\left(\frac{\pi_\theta(o_i | q)}{\pi_{\theta_{\text{old}}}(o_i | q)}, 1 \pm \varepsilon\right) A_i \right) - \beta \text{KL}[\pi_\theta \| \pi_{\text{ref}}] \right]
\label{eq-6}
\end{equation}
where $A_i = \frac{r_i - \text{mean}(r_1, r_2, \ldots, r_G)}{\text{std}(r_1, r_2, \ldots, r_G)}$ represents the advantage function computed through group-based calculation, $r_i$ denotes the reward value for answer $o_i$, $\text{KL}[\pi_{\theta}||\pi_\mathrm{ref}]$ represents the KL divergence~\citep{kullback1951kl} between the policy model and reference model, and $\beta$ is the regularization hyperparameter.

%% file: sections/5.experiments.tex
\section{Experiments}

\subsection{Experimental Setup}

\paragraph{Implementation Details.} 
We implement \methodname{} using Qwen2.5-VL-3B~\citep{bai2025qwen2} as the foundation model. The training procedure follows a three-stage progressive schedule with stage-specific hyperparameter configurations. Stages 1 and 2 employ supervised fine-tuning~\citep{ouyang2022sft}, while Stage 3 utilizes GRPO~\cite{guo2025deepseek} for reinforcement learning. Additional training details are provided in~\ref{app:training_implementation}.

\paragraph{Evaluation Benchmarks.} 
We evaluate \methodname{} on six benchmarks across in-domain and out-of-domain settings. For in-domain evaluation, we use VSI-Bench~\citep{yang2025thinking} containing 5,155 video-based spatial reasoning questions and introduce two new benchmarks: SPBench-SI (1,009 single-image questions) and SPBench-MV (319 multi-view questions). Both SPBench benchmarks are constructed from ScanNet validation scenes using our pipeline, with strict scene-level separation ensuring zero overlap with training data. For out-of-domain evaluation, we assess generalization on CV-Bench~\citep{tong2024cambrian} for 2D/3D vision tasks, SPAR-Bench~\citep{zhang2025flatland} for multi-difficulty spatial reasoning, and ViewSpatial-Bench~\citep{li2025viewspatial} for perspective-dependent spatial understanding. Detailed benchmark and baseline descriptions are provided in Appendices~\ref{app:details_of_benchmarks} and~\ref{app:details_of_baselines}, respectively.

\subsection{Main Results}

\paragraph{In-domain Performance.} Table \ref{tab:in-domain} presents comprehensive evaluation on spatial reasoning benchmarks. \methodname{} achieves state-of-the-art performance with 62.3\% overall accuracy, surpassing all baselines including proprietary models. The performance gain is particularly pronounced on our proposed benchmarks: 70.2\% on SPBench-SI (+29.9\% over base model) and 70.9\% on SPBench-MV (+34.3\% over base model), demonstrating the effectiveness of our progressive training approach.
Notably, while Spatial-MLLM achieves competitive performance on VSI-Bench (47.3\%) using specialized 3D encoders, \methodname{} attains comparable results of 45.7\% (+16.3\% over base model) using only the standard VLM architecture, validating that progressive training can substitute for architectural modifications. The consistent improvements across both numerical questions and multiple-choice questions indicate robust spatial understanding rather than task-specific overfitting. Further details of in-domain performance are presented in~\ref{app:in-domain_details}.

\paragraph{Generalization Analysis.} 
Table \ref{tab:out-of-domain} demonstrates strong out-of-domain generalization with 50.8\% overall accuracy, outperforming GPT-4o (48.1\%) and maintaining 7.2\% improvement over the base model. The gains are consistent across diverse evaluation settings: CV-Bench tests classical vision tasks, SPAR-Bench evaluates multi-difficulty reasoning, and ViewSpatial-Bench assesses perspective-dependent understanding. Particularly noteworthy is the 16.5\% improvement on person-perspective tasks in ViewSpatial-Bench, suggesting that our training develops robust spatial representations that transfer to novel viewpoints. Further details of out-of-domain performance are presented in~\ref{app:out-of-domain_details}.

\begin{table}[t]
\centering
\small
\caption{\textbf{Evaluation Results on In-domain Benchmarks.} NQ and MCQ denotes numerical question and multiple-choice question, respectively. For each metric, \textbf{bold} numbers indicate the best performance, while \underline{underlined} numbers represent the second-best performance.}
\resizebox{\textwidth}{!}{  %
\begin{tabular}
{lC{0.8cm}C{0.8cm}C{0.8cm}C{0.8cm}C{0.8cm}C{0.8cm}C{0.8cm}C{0.8cm}C{0.8cm}C{1.2cm}}
\toprule
\multirow{2}{*}{\textbf{Model}} & \multicolumn{3}{c}{\textbf{VSI-Bench}} & \multicolumn{3}{c}{\textbf{SPBench-SI}} & \multicolumn{3}{c}{\textbf{SPBench-MV}} & \multirow{2}{*}{\textbf{Overall}} \\
\cmidrule(lr){2-4} \cmidrule(lr){5-7} \cmidrule(lr){8-10}
& NQ & MCQ & Avg. & NQ & MCQ & Avg. & NQ & MCQ & Avg.& \\
\midrule
\rowcolor{gray!10} 
\multicolumn{11}{l}{\textit{\textbf{Proprietary Models}}} \\
GPT-4o~\citep{hurst2024gpt} & 33.4 & 34.6 & 34.0 & 24.5 & 60.3 & 42.4 & 40.7 & 59.4 & 48.2 & 41.5 \\
Gemini-2.0-Flash~\citep{team2024gemini} & 46.4 & \textbf{44.3} & 45.4 & \underline{49.0} & 60.4 & \underline{54.7} & 51.9 & 50.7 & 56.5 & \underline{52.2} \\
\midrule
\rowcolor{gray!10} 
\multicolumn{11}{l}{\textit{\textbf{Open-Source Models}}} \\
InternVL-2.5-4B~\citep{chen2024internvl} & 30.6 & 34.1 & 32.6 & 31.8 & 53.3 & 42.5 & 37.7 & 51.4 & 43.2 & 42.8 \\
InternVL-2.5-8B~\citep{chen2024internvl} & 40.4 & 40.0 & 40.2 & 28.3 & 56.3 & 42.3 & 37.3 & 47.5 & 41.4 & 41.4 \\
Kimi-VL-A3B~\citep{team2025kimi} & 31.8 & 25.5 & 28.7 & 25.7 & 44.9 & 35.3 & 23.3 & 57.6 & 37.0 & 36.0 \\
LLaVA-OneVision-7B~\citep{li2024llava-one} & 34.5 & 31.2 & 33.1 & 25.4 & 41.0 & 33.2 & 20.6 & 49.6 & 32.2 & 32.2 \\
\midrule
\rowcolor{gray!10} 
\multicolumn{11}{l}{\textit{\textbf{Qwen2.5-VL-7B Based Spatial Models}}} \\
Qwen2.5-VL-7B~\citep{bai2025qwen2} & 37.1 & 34.6 & 35.8 & 36.3 & 60.5 & 48.4 & 28.9 & 49.8 & 37.3 & 43.9 \\
SpaceR-7B~\citep{ouyang2025spacer} & 47.8 & 41.2 & 44.5 & 35.7 & 61.5 & 48.6 & 63.2 & 53.7 & \underline{59.4} & 50.8 \\
VILASR-7B~\citep{wu2025reinforcing} & 47.4 & \underline{43.4} & 45.4 & 36.6 & \underline{63.7} & 50.2 & 56.2 & \underline{59.6} & 57.6 & 51.1 \\
Video-R1~\citep{feng2025video} & 33.8 & 32.9 & 33.4 & 27.7 & 62.0 & 44.8 & 32.5 & 53.0 & 40.7 & 39.6 \\
\midrule
\rowcolor{gray!10} 
\multicolumn{11}{l}{\textit{\textbf{Qwen2.5-VL-3B Based Spatial Models}}} \\
Qwen2.5-VL-3B~\citep{bai2025qwen2} & 26.0 & 33.0 & 29.4 & 24.3 & 56.2 & 40.3 & 25.6 & 53.2 & 36.6 & 38.8 \\
Spatial-MLLM-4B~\citep{wu2025spatial} & \textbf{51.5} & 43.1 & \textbf{47.3} & 38.1 & 49.3 & 43.7 & \underline{63.7} & 58.9 & 53.1 & 48.0 \\
\methodname-3B & \underline{50.8} & 40.5 & \underline{45.7} & \textbf{58.6} & \textbf{81.8} & \textbf{70.2} & \textbf{68.2} & \textbf{75.0} & \textbf{70.9} & \textbf{62.3} \\
\textcolor{ForestGreen}{\textit{\textbf{Improvement}}} & \textcolor{ForestGreen}{+24.9} & \textcolor{ForestGreen}{+7.6} & \textcolor{ForestGreen}{+16.3} & \textcolor{ForestGreen}{+34.3} & \textcolor{ForestGreen}{+25.6} & \textcolor{ForestGreen}{+29.9} & \textcolor{ForestGreen}{+42.6} & \textcolor{ForestGreen}{+21.8} & \textcolor{ForestGreen}{+34.3} & \textcolor{ForestGreen}{+23.4}\\
\bottomrule
\end{tabular}
}
\label{tab:in-domain}
\vspace{-10pt}
\end{table}

\begin{table}[t]
\centering
\small
\caption{\textbf{Evaluation Results on Out-of-domain Benchmarks.} For ViewSpatial-Bench, CP and PP represent Camera-Perspective and Person-Perspective, respectively. For each metric, \textbf{bold} numbers indicate the best performance, while \underline{underlined} numbers represent the second-best performance.}
\resizebox{\textwidth}{!}{
\begin{tabular}{lC{0.7cm}C{0.7cm}C{0.7cm}C{0.7cm}C{1.2cm}C{0.7cm}C{0.7cm}C{0.7cm}C{0.7cm}C{0.7cm}C{0.9cm}}
\toprule
\multirow{2}{*}{\textbf{Model}} & \multicolumn{3}{c}{\textbf{\mbox{CV-Bench}}} & \multicolumn{4}{c}{\textbf{SPAR-Bench}} & \multicolumn{3}{c}{\textbf{\mbox{ViewSpatial-Bench}}} & \multirow{2}{*}{\textbf{Overall}} \\
\cmidrule(lr){2-4} \cmidrule(lr){5-8}\cmidrule(lr){9-11}
& 2D & 3D & Avg. & Low & Medium & High & Avg. &CC & PC & Avg. & \\
\midrule
GPT-4o & 69.4 & \underline{81.3} & \underline{75.4} & \textbf{29.3} & 24.9 & \textbf{45.1} & \textbf{36.4} & 33.7 & 31.5 & 32.6 & 48.1 \\
InternVL-2.5-4B & \underline{73.5} & 75.1 & 74.4 & 25.7 & \underline{29.8} & 35.2 & 30.6 & \underline{40.8} & 35.1 & \underline{37.9} & 47.6 \\
Kimi-VL-A3B & 41.9 & 54.7 & 48.3 & \underline{26.4} & 18.8 & 36.2 & 29.4 & 25.1 & \underline{41.5} & 33.6 & 37.1 \\
LLaVA-OneVision-7B & 53.2 & 63.5 & 58.3 & 21.8 & 26.1 & 40.1 & 31.2 & 28.5 & 26.5 & 27.5 & 39.0 \\
Qwen2.5-VL-7B & \textbf{75.0} & \textbf{83.1} & \textbf{79.0} & 17.5 & 29.5 & 41.8 & 30.2 & \textbf{41.8} & 34.3 & 37.9 & \underline{49.0} \\
\midrule
\rowcolor{gray!10}
Qwen2.5-VL-3B & 69.1 & 72.2 & 70.6 & 15.3 & 26.4 & 32.2 & 24.6 & 39.5 & 32.0 & 35.6 & 43.6 \\
\rowcolor{gray!10}
\methodname-3B & 72.4 & 74.9 & 73.7 & 25.8 & \textbf{33.2} & \underline{42.6} & \underline{34.4} & 39.6 & \textbf{48.5} & \textbf{44.2} & \textbf{50.8} \\
\rowcolor{gray!10}
\textcolor{ForestGreen}{\textit{\textbf{Improvement}}} & \textcolor{ForestGreen}{+3.3} & \textcolor{ForestGreen}{+2.7} & \textcolor{ForestGreen}{+3.1} & \textcolor{ForestGreen}{+10.5} & \textcolor{ForestGreen}{+6.8} & \textcolor{ForestGreen}{+10.4} & \textcolor{ForestGreen}{+9.8} & \textcolor{ForestGreen}{+0.1} & \textcolor{ForestGreen}{+16.5} & \textcolor{ForestGreen}{+8.9} & \textcolor{ForestGreen}{+7.2} \\
\bottomrule
\end{tabular}
}
\label{tab:out-of-domain}
\vspace{-10pt}
\end{table}

\subsection{Ablation Studies}

\paragraph{Component Analysis.}
Figure \ref{fig:ablation_study} reveals the critical interdependence of \methodname's components. Stage 2 (spatial understanding fine-tuning) proves most essential, with its removal causing a 9.4\% accuracy drop, validating explicit spatial cognition as the training cornerstone. Stages 1 and 3 contribute meaningfully (1.8\% and 2.1\% drops respectively), confirming progressive training's value. Excluding single-image and multi-view data causes the most severe degradation (16.4\% loss), affecting not only corresponding benchmarks but also video-based VSI-Bench performance. This demonstrates that multimodal diversity is fundamental for robust spatial reasoning across all modalities. Chain-of-thought reasoning provides consistent 0.8\% gains, validating explicit reasoning in spatial tasks.

\begin{figure}[t]
	\centering
	\includegraphics[width=1\linewidth]{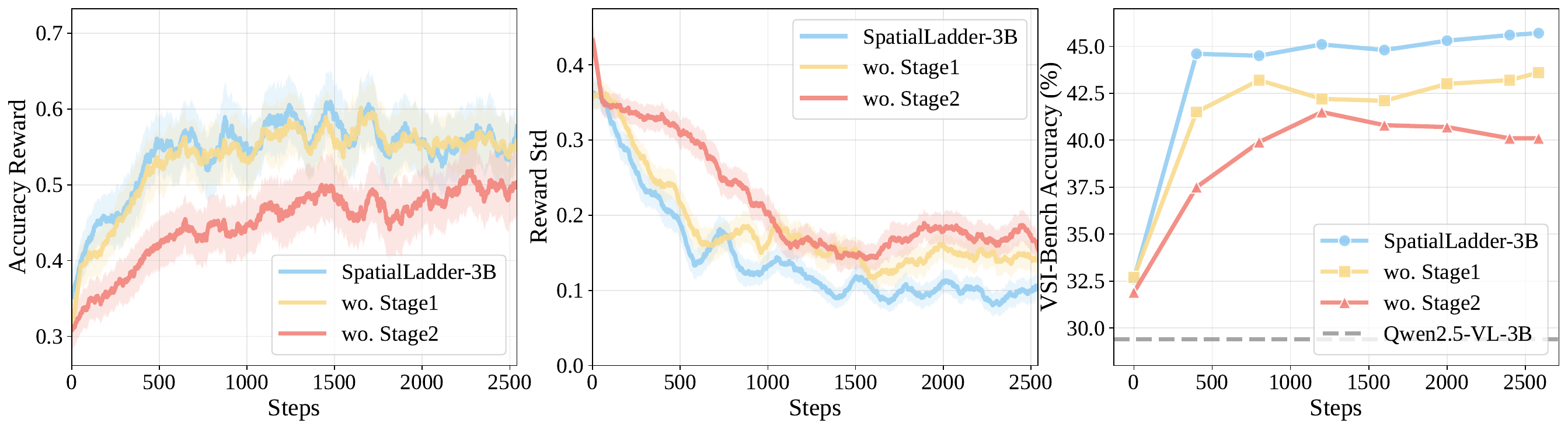}
	\caption{\textbf{Impact of progressive training stages.} Left: accuracy rewards over training steps; Middle: reward standard deviation over training steps; Right: VSI-Bench performance comparison.}
	\label{fig:stage_dynamics}
    \vspace{-10pt}
\end{figure}

\paragraph{Training Dynamics.}
Figure \ref{fig:stage_dynamics} demonstrates that the complete \methodname-3B consistently outperforms variants missing Stage 1 or Stage 2 across accuracy reward curves. The reward standard deviation analysis reveals superior training stability for the full model, exhibiting the most significant variance reduction and smoothest convergence patterns. On VSI-Bench evaluation, the complete framework achieves highest accuracy while ablated variants show notable degradation, with Stage 2's absence producing the most pronounced performance decline. Appendix~\ref{app:cot_dynamics} provides additional training dynamics with and without chain-of-thought.

\subsection{In-depth Analysis}

\begin{figure}[t]
  \centering
  \makebox[\textwidth][c]{%
      \begin{minipage}{0.45\textwidth}
        \centering
        \includegraphics[height=3.8cm]{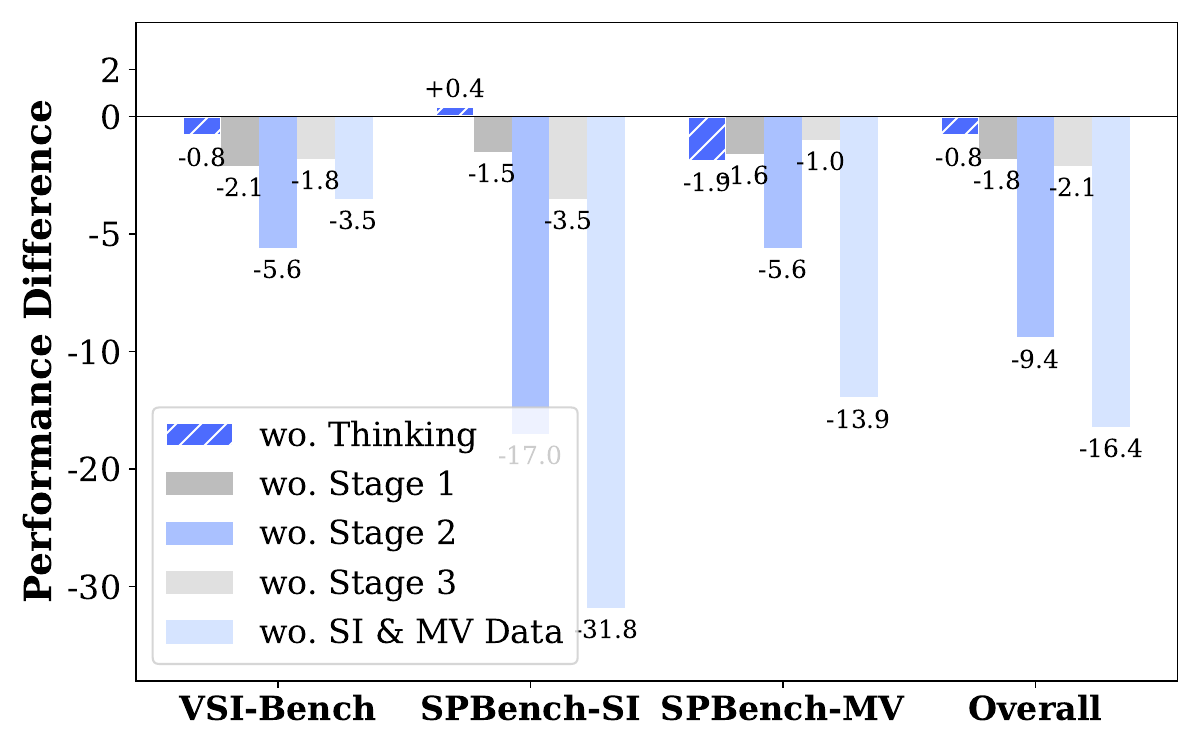}
        \caption{Ablation study results.}
        \label{fig:ablation_study}
      \end{minipage}
      \hfill
      \begin{minipage}{0.54\textwidth}
        \centering
        \includegraphics[height=3.8cm]{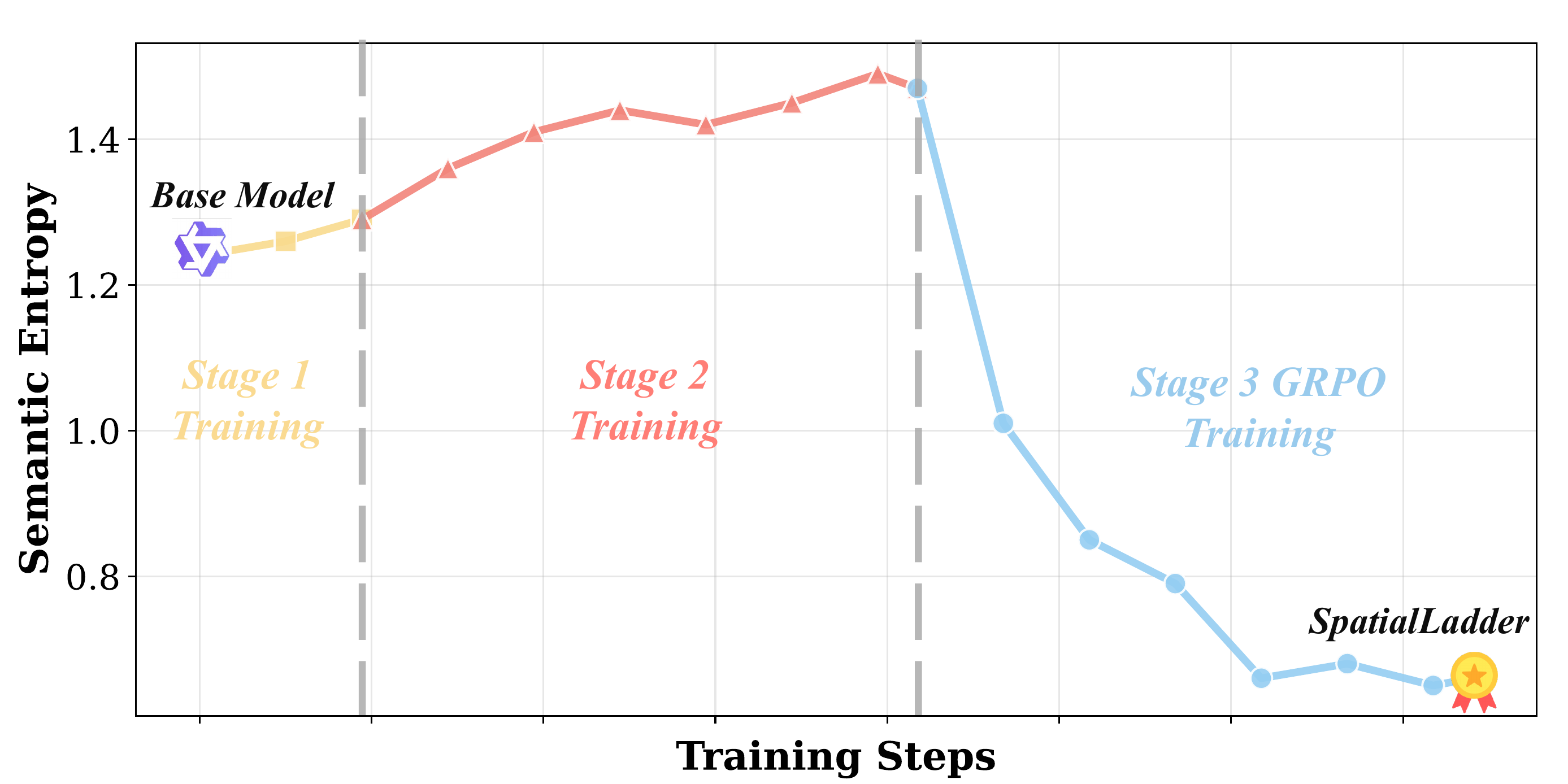}
        \caption{Semantic entropy dynamics.}
        \label{fig:semantic_entropy_dynamics}
      \end{minipage}
  }
  \vspace{-10pt}
\end{figure}

\paragraph{Semantic Consistency Emerges through Reinforcement Optimization.}
We employ semantic entropy~\citep{kuhn2023semantic} to quantify model uncertainty. As shown in Figure~\ref{fig:semantic_entropy_dynamics}, during Stages 1-2 where the model establishes perceptual foundations and spatial understanding capabilities, entropy increases from 1.24 to 1.47 as spatial capabilities transcend initial misconceptions and expand the exploration space for improved reasoning. Subsequently, during Stage 3 reinforcement learning, entropy steadily declines from 1.47 to 0.66, marking the transition from broad exploration to focused reasoning convergence. This quantitative progression validates our three-stage strategy: establishing comprehensive foundations, expanding the reasoning space, and achieving robust spatial intelligence through convergence. Further details are provided in Appendix~\ref{app:entropy_details}.

\begin{figure}[t]
	\centering
	\includegraphics[width=1\linewidth]{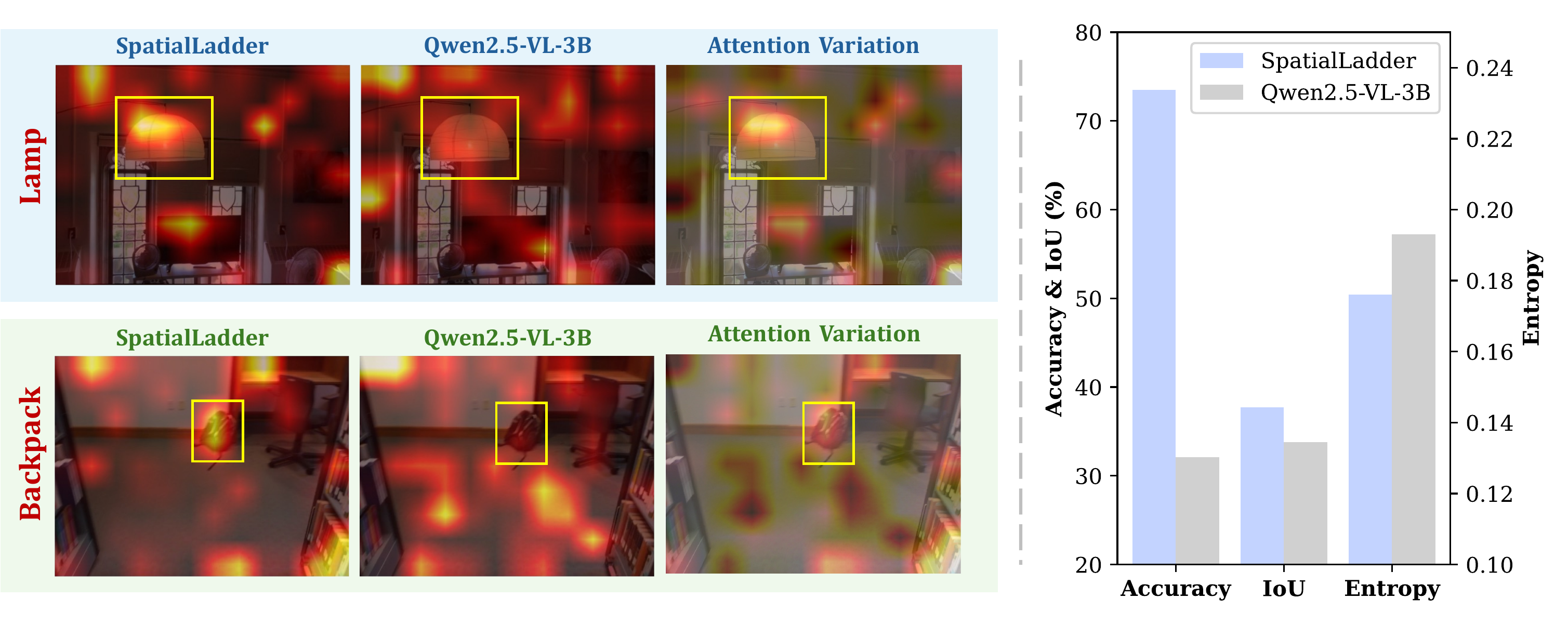}
	\caption{\textbf{Visual attention comparison between \methodname{} and Qwen2.5-VL-3B.} Left: Representative attention distribution patterns for both models. Right: Quantitative analysis of performance accuracy, attention IoU, and attention entropy metrics.}
	\label{fig:attention}
    \vspace{-10pt}
\end{figure}

\paragraph{Visual Attention Becomes Precisely Object-centric through Progressive Training.}
To understand how our training framework influences internal mechanisms, we analyzed visual attention patterns of \methodname{} and Qwen2.5-VL-3B during object size estimation tasks. Figure~\ref{fig:attention} (left) reveals \methodname{} exhibits significantly more concentrated attention on task-relevant objects.

We conducted a quantitative evaluation of attention distributions using 400 samples from SPBench-SI, with two metrics: Visual Attention IoU, which measures the concentration of attention within object bounding boxes, and Visual Attention Entropy, which quantifies the degree of attention dispersion across the visual field. Results in Figure 7 (right) demonstrate that SpatialLadder achieves superior performance with 73.5\% accuracy and 37.7\% visual attention IoU compared to the base model's 32.1\% accuracy and 33.8\% IoU. Additionally, SpatialLadder exhibits lower attention entropy (0.176 vs. 0.193), indicating more concentrated focus on relevant objects. These quantitative findings provide empirical evidence that our training framework significantly enhances precise spatial object perception and spatial understanding. Further details about visual attention analysis are provided in~\ref{app:attention_details}.

\begin{figure}[t]
	\centering
	\includegraphics[width=1\linewidth]{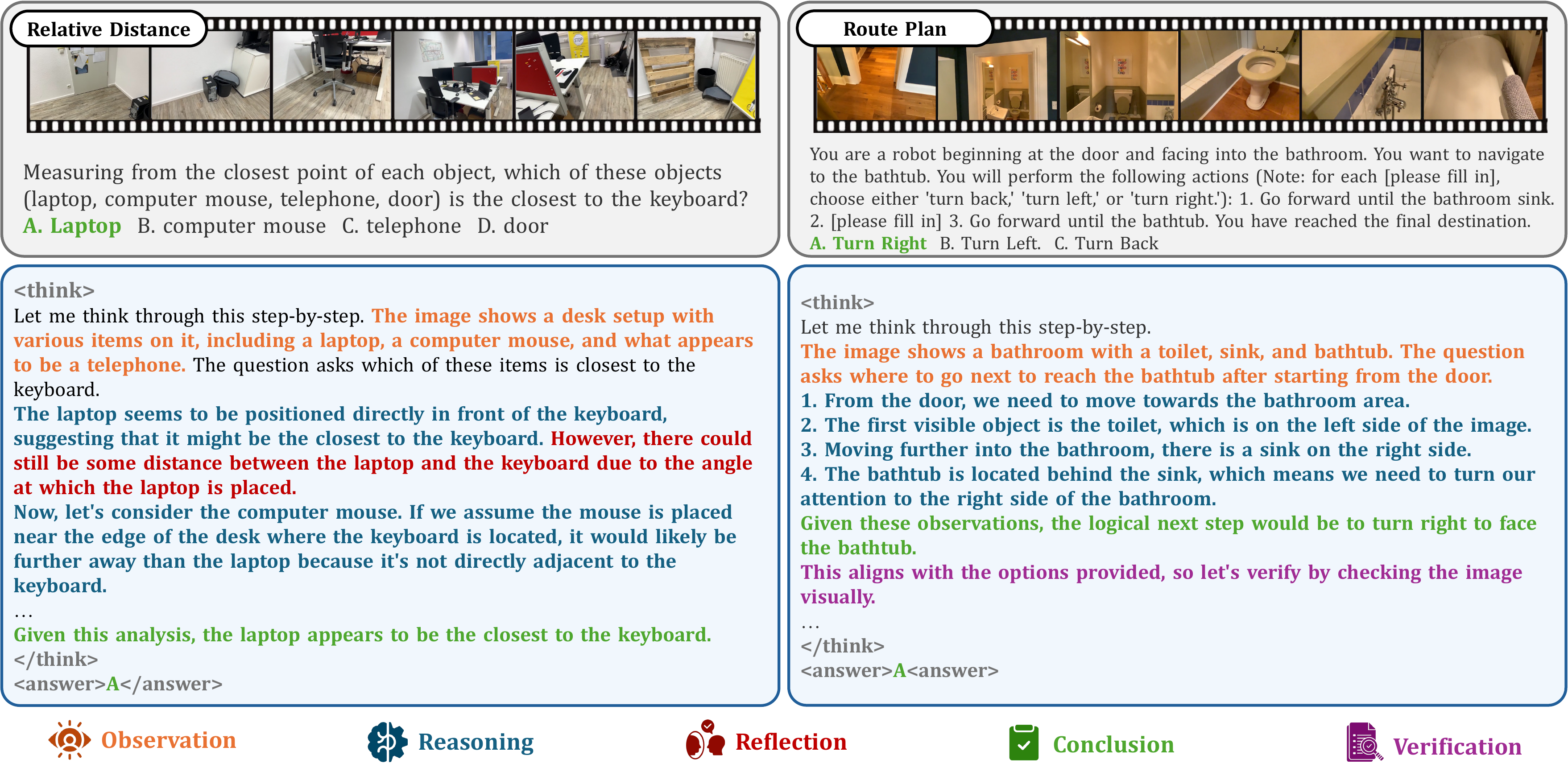}
	\caption{Hierarchical reasoning process demonstration in \methodname.}
	\label{fig:cases}
    \vspace{-10pt}
\end{figure}

\paragraph{Hierarchical reasoning structures develop naturally from perceptual foundations.}
Qualitative analysis reveals that \methodname{} develops systematic spatial cognition from foundational perception training. Figure~\ref{fig:cases} demonstrates a hierarchical cognitive architecture: accurate spatial element identification provides the perceptual foundation for constructing logical reasoning chains through structured analysis. The model exhibits sophisticated metacognitive capabilities, including self-verification and error correction mechanisms ensuring reasoning consistency.

In relative distance tasks, \methodname{} systematically decomposes spatial relationships, while path-planning scenarios show structured consideration of complex layouts. This chain-of-thought mechanism demonstrates how perceptual foundations naturally scaffold higher-order reasoning abilities. 
Importantly, SpatialLadder delivers correct conclusions with clear reasoning, showing that foundational spatial perception supports hierarchical reasoning and validating the effectiveness of our training framework.

Additional analysis about comparison with other spatial reasoning dataset, dataset scaling and progressive training order are provided in~\ref{app:spatial_data_comparison},~\ref{app:dataset_scaling} and~\ref{app:progressive} respectively.

%% file: sections/6.conclusion.tex
\section{Conclusion}

This work addresses the perception–reasoning gap in VLMs for spatial tasks by proposing a systematic solution. We introduce \datasetname, a multimodal dataset covering object localization, single-view, multi-view, and video-based spatial reasoning. We design a three-stage progressive training framework that builds spatial intelligence from perception to understanding and reasoning, effectively bridging this gap. Our \methodname{} model achieves state-of-the-art results on multiple benchmarks, demonstrating strong in-domain and out-of-domain performance. Ablation studies confirm the effectiveness of each component. This approach establishes a new paradigm for spatial reasoning in VLMs, opening promising directions for future research as discussed in Appendix~\ref{app:limitations}.

%% file: sections/8.statement.tex
\newpage

\section*{Ethics Statement}

This work does not involve human subjects, personal data, or sensitive information. All datasets used in our experiments (VSI-Bench, SPBench-SI, SPBench-MV, CV-Bench, SPAR-Bench, ViewSpatial-Bench) are publicly available benchmark datasets designed for evaluating visual spatial reasoning in VLMs. We strictly adhered to ethical research practices and did not conduct any data collection that could raise privacy, security, or fairness concerns. Our methods—\datasetname{} dataset and progressive three-stage training framework—address the perception-reasoning gap in VLMs for spatial tasks, developing robust spatial reasoning capabilities without introducing risks of harmful applications. To the best of our knowledge, this research complies with the ICLR Code of Ethics and poses no foreseeable ethical concerns.

\section*{Reproducibility Statement}

We have made extensive efforts to ensure the reproducibility of our work. Comprehensive details of dataset construction are provided in~\ref{app:dataset_construction}, while training configurations and hyperparameters are systematically reported in~\ref{app:training_implementation}. Detailed dataset descriptions are documented in~\ref{app:details_of_benchmarks}. The comprehensive evaluation results are outlined in~\ref{app:in-domain_details} and~\ref{app:out-of-domain_details}, and the implementation details of our analysis experiments are thoroughly described in~\ref{app:entropy_details} and~\ref{app:attention_details}. Upon acceptance, we will release our models, together with training and evaluation code, to facilitate replication and further research.

%% file: sections/9.bib.tex
\bibliographystyle{iclr2026_conference}
\bibliography{ref}

%% file: sections/99.appendix.tex
\appendix
\newpage
\section*{Technical Appendices and Supplementary Material}

\section{Preliminary Analysis}
\label{app:preliminary}

To validate our hypothesis that spatial reasoning failures stem from inadequate perceptual grounding rather than reasoning incapacity, we conducted controlled experiments examining how progressive perceptual hints affect model performance. We constructed a diagnostic dataset of 200 spatial orientation tasks from ScanNet validation scenes, each requiring determination of relative positions between object pairs from the camera's perspective.

We evaluated Qwen2.5-VL-3B~\citep{bai2025qwen2} under three conditions: (1) baseline with raw images and questions, (2) location hints adding colored bounding boxes around target objects, and (3) full hints incorporating directional arrows within bounding boxes. Results demonstrate monotonic improvement with enhanced perceptual grounding: baseline accuracy of 36.5\% improves to 41.5\% with location hints (+5.0\%) and 46.0\% with directional hints (+4.5\% additional).

These findings directly motivate our progressive training approach. This experimental evidence establishes that robust spatial reasoning cannot be achieved through end-to-end learning but requires systematic construction from perceptual foundations to abstract reasoning. After training, our model \methodname{} achieves consistently high performance across all conditions: 82.0\% without hints, 82.5\% with location hints, and 83.5\% with full hints. This minimal variation (1.5\% range) demonstrates that progressive training successfully internalizes spatial perception capabilities, eliminating dependence on external scaffolding.

\section{Additional Method Details}

\subsection{Details of \datasetname{} Construction}
\label{app:dataset_construction}

\begin{table}[h]
\caption{Question templates for tasks in \datasetname.}
\centering
\label{tab:question_template}
\resizebox{\linewidth}{!}{
\begin{tabular}{r | p{12cm}}
\toprule
\textbf{Task} & \textbf{Question Template} \\
\midrule
Object Counting & \textit{How many \textbf{\{category\}}(s) appear?} \\
\midrule
Absolute Distance & \textit{Measuring from the closest point of each object, what is the distance between the \textbf{\{object 1\}} and the \textbf{\{object2\}} (in meters)?} \\
\midrule
Object Size & \textit{What is the length of the longest dimension (length, width, or height) of the \textbf{\{object\}}, measured in centimeters?} \\
\midrule
Relative Distance & \textit{Measuring from the closest point of each object, which of these two objects (\textbf{\{choice a\}}, \textbf{\{choice b\}}) is closer to the \textbf{\{category\}}?} \\
\midrule
Relative Direction & \parbox{12cm}{
The question template for relative direction tasks varies between single-image and multi-view modalities.
\begin{itemize}[label=\textbullet, leftmargin=*]
    \item \textbf{Single-image}: \textit{From the camera's perspective, is the \textbf{\{object 1\}} to the \textbf{\{object 2\}}'s \textbf{\{choice a\}}, \textbf{\{choice b\}}, \textbf{\{choice c\}} or \textbf{\{choice d\}}?}
    \item \textbf{Multi-view}: \textit{If I am standing by the \textbf{\{positioning object\}} and facing the \textbf{\{orienting object\}}, is the \textbf{\{querying object\}} to my \textbf{\{choice a\}}, \textbf{\{choice b\}}, \textbf{\{choice c\}} or \textbf{\{choice d\}}? The directions refer to the quadrants of a Cartesian plane (if I am standing at the origin and facing along the positive y-axis).?}
\end{itemize}
} \\
\midrule
Object Localization & \textit{\textbf{\{question\}} Please carefully observe the image first to identify the object(s) referred to in the question. Note that each object type appears only once in the image. Then provide the 2D bounding box coordinates and labels of the related objects in JSON format.} \\ 
\bottomrule
\end{tabular}
}
\vspace{-10pt}
\end{table}

\paragraph{Question-answer Generation Details}
Based on the metadata unified from ScanNet, we further construct question–answer pairs for various single-image and multi-view spatial reasoning tasks. The generation process is described as follows:
\begin{itemize}[label=\textbullet]
    \item \textbf{Object Counting}: This task involves a single object category and is designed exclusively for the multi-view setting. We first identify the target object category, then determine the number of distinct instances that appear across all views by checking their presence in each camera view, and finally use the aggregated count as the answer.
    \item \textbf{Absolute Distance}: This task involves two objects. Using their 3D locations from the metadata, we calculate the Euclidean distance between them as the answer. To ensure clarity and avoid ambiguity, we enforce that the computed distance must exceed the minimum size of the two objects.
    \item \textbf{Object Size}: This task focuses on a single object. We estimate the object’s size using the maximum dimension of its 3D bounding box, while filtering out objects that are either excessively large or small to ensure human-scale spatial reasoning.
    \item \textbf{Relative Distance}: This task involves three objects—a target object and two candidate objects. We compare the distances from the target object to each candidate and select the closer one as the answer. To prevent ambiguous cases, we require the larger distance to be at least twice the smaller one.
    \item \textbf{Relative Direction}: The construction of this task differs between single-image and multi-view modalities. For the single-image setting, the task involves two objects, and their relative orientation is defined with respect to the camera viewpoint. Specifically, we compute the left/right relation based on their 2D locations relative to the image plane, and the front/back relation based on their depth from the camera. Composite relations (e.g., left-front, left-back, right-front, right-back) are included when applicable. For the multi-view setting, the task involves three objects: a positioning object, an orienting object, and a querying object. We define vectors from the positioning object to the orienting object ($\vec{a}$) and to the querying object ($\vec{b}$), and compute the angle $\theta$ between them using $\cos(\theta)=\frac{\vec{a}\cdot\vec{b}}{|\vec{a}||\vec{b}|}$. The relative direction of the querying object is then determined according to this angular relationship.
\end{itemize}
The question templates employed for generating QA pairs across various spatial reasoning and object localization tasks in \datasetname{} are detailed in Table \ref{tab:question_template}.

\paragraph{Quality Assurance.}
To ensure dataset reliability, we implement multiple filtering mechanisms. Scene diversity is maintained by limiting questions per scene to prevent overfitting to specific environments. Object diversity is enforced by restricting the number of samples constructed from the same object type within a scene, avoiding bias toward particular categories and ensuring question variety. Noisy objects (e.g. wall, floor and ceiling) are filtered to focus on human-scale spatial reasoning. Minimum visibility threshold (40\%) ensures that spatial judgments are based on sufficient visual evidence. Objects must be uniquely identifiable within their context to avoid ambiguity. These constraints eliminate approximately 90\% of initially generated samples, resulting in a high-quality dataset where each sample provides clear spatial learning signals.

\begin{table}[h]
\caption{Detailed statistics of the spatial reasoning subset in \datasetname.}
\centering
\label{tab:dataset_statics}
\resizebox{\linewidth}{!}{
\begin{tabular}{lcccccccc}
\toprule
\multirow{2}{*}{\textbf{Modality}} & \multicolumn{4}{c}{\textbf{Numerical Question}} & \multicolumn{3}{c}{\textbf{Multiple-choice Question}} & \multirow{2}{*}{\textbf{Total}} \\
\cmidrule(r){2-5} \cmidrule(l){6-8}
& Obj. Cnt. & Abs. Dist. & Obj. Size & Room Size & Rel. Dist. & Rel. Dir. & Appr. Order & \\
\midrule
Single-Image & - & 1,127 & 1,514 & - & 1,034 & 2,253 & - & 5,929 \\
Multi-View & 217 & 817 & 1,867 & - & 635 & 2,162 & - & 5,752 \\
Video & 507 & 1,500 & 1,331 & 150 & 1,134 & 3,061 & 1,317 & 9,000 \\
\bottomrule
\end{tabular}
}
\vspace{-10pt}
\end{table}

\paragraph{The Statics of \datasetname{}.} The distribution of spatial reasoning tasks across different modalities in our constructed \datasetname{} is presented in Table~\ref{tab:dataset_statics}. In \datasetname{}, object localization tasks are based on the single-image modality but remain independent of specific spatial reasoning task types. Each single-image spatial reasoning task is paired with a corresponding object localization task.

\subsection{Details of Training Implementation}
\label{app:training_implementation}

\paragraph{Prompt Used for Training.} The system prompt and user prompt employed in the \methodname{} three-stage training framework are presented in the boxes below. The post prompt design in Stage 2 and Stage 3 varies across task types: for multiple-choice questions, the prompt guides the model to output the corresponding option, whereas for numerical questions, the prompt instructs the model to provide a numerical answer.

\begin{AIbox}{Prompt for Stage 1}
\small
    \textbf{System Prompt}: \textit{``You are a helpful assistant.''} \\
    \textbf{User Prompt}: \textit{\textbf{\{question\}} + ``Please carefully observe the image first to identify the object(s) referred to in the question. Note that each object type appears only once in the image. Then provide the 2D bounding box coordinates and labels of the related objects in JSON format.''}
\end{AIbox}

\begin{AIbox}{Prompt for Stage 2}
\small
    \textbf{System Prompt}: \textit{``You are a helpful assistant.''} \\
    \textbf{User Prompt}: \textit{\textbf{\{question\}}} + \textbf{Post Prompt[\textit{``question type''}]} \\
    \textbf{Post Prompt}:
    \begin{itemize}[label=\textbullet, leftmargin=*]
        \item Multiple-choice Question: \textit{``Please answer with the option's letter from the given choices (e.g., A, B, etc.) directly.''}
        \item Numerical Question: \textit{``Please answer the question using a numerical value (e.g., 42 or 3.1) directly.''}
    \end{itemize}
\end{AIbox}

\begin{AIbox}{Prompt for Stage 3}
\small
    \textbf{System Prompt}: \textit{``You are a helpful assistant.''} \\
    \textbf{User Prompt}: \textit{\textbf{\{question\}}} + \textit{``Please think about this question as if you were a human pondering deeply. Engage in an internal dialogue using expressions such as 'let me think', 'wait', 'Hmm', 'oh, I see', 'let's break it down', etc, or other natural language thought expressions. It's encouraged to include self-reflection or verification in the reasoning process.''} + \textbf{Post Prompt[\textit{``question type''}]} \\
    \textbf{Post Prompt}:
    \begin{itemize}[label=\textbullet, leftmargin=*]
        \item Multiple-choice Question: \textit{``Please provide your detailed reasoning between the \texttt{<think>} \texttt{</think>} tags, and then answer the question with the option's letter from the given choices (e.g., A, B, etc.) within the \texttt{<answer>} \texttt{</answer>} tags."''}
        \item Numerical Question: \textit{``Please provide your detailed reasoning between the \texttt{<think>} \texttt{</think>} tags, and then answer the question with a numerical value (e.g., 42 or 3.1) within the \texttt{<answer>} \texttt{</answer>} tags.''}
    \end{itemize}
\end{AIbox}

\begin{table}[H]
  \small
  \centering
  \makebox[\textwidth][c]{%
      \begin{minipage}{0.48\linewidth}
        \centering
        \caption{Hyperparameter used in Stage 1-2.}
        \begin{tabular}{ll}
            \toprule
            \textbf{Hyperparameter}            & \textbf{Value}      \\ \midrule
            per\_device\_train\_batch\_size    & 1                   \\
            gradient\_accumulation\_steps      & 8                   \\
            bf16                               & true                \\
            data\_seed                         & 42                  \\ 
            gradient\_checkpointing            & true                \\
            attn\_implementation               & flash\_attention\_2 \\
            lr\_scheduler\_type                & cosine              \\
            warmup\_ratio                      & 0.1                 \\
            num\_train\_epochs                 & 1                   \\
            max\_pixels                        & 100,352             \\
            min\_pixels                        & 12,544              \\ \bottomrule
            \end{tabular}
        \label{tab:hyper_stage12}
      \end{minipage}
      \hfill
      \begin{minipage}{0.48\linewidth}
        \centering
        \caption{Hyperparameter used in Stage 3.}
        \begin{tabular}{ll}
            \toprule
            \textbf{Hyperparameter}            & \textbf{Value}      \\ \midrule
            num\_generations                   & 8                   \\
            per\_device\_train\_batch\_size    & 2                   \\
            gradient\_accumulation\_steps      & 4                   \\
            bf16                               & true                \\
            data\_seed                         & 42                  \\ 
            gradient\_checkpointing            & true                \\
            attn\_implementation               & flash\_attention\_2 \\
            num\_train\_epochs                 & 1                 \\
            max\_pixels                        & 100,352           \\
            min\_pixels                        & 12,544            \\
            $\beta$                            & 0.01              \\ \bottomrule
            \end{tabular}
        \label{tab:hyper_stage3}
      \end{minipage}
    }
  \vspace{-10pt}
\end{table}

\paragraph{Reproduction details.}

Our model was trained on a 4 × NVIDIA A6000 GPU cluster with 48GB memory per device. The training process consisted of three distinct stages: stages 1-2 employed supervised fine-tuning methodology implemented via the HuggingFace Transformers Reinforcement Learning (TRL) framework, with corresponding hyperparameters detailed in Table~\ref{tab:hyper_stage12}. Stage 3 utilized GRPO reinforcement learning, implemented through the VLM-R1 framework~\citep{shen2025vlm-r1}, with corresponding hyperparameters specified in Table~\ref{tab:hyper_stage3}.

\paragraph{Details of Stage 1 and Stage 2.}
In our progressive three-stage training framework, Stage 1 develops spatial perception capabilities through targeted spatial localization tasks, while Stage 2 enhances spatial understanding through multi-dimensional spatial reasoning tasks. Both stages employ supervised fine-tuning methodology, optimizing the standard cross-entropy loss function:

\begin{equation}
\mathcal{L}_{\text{ce}}(\theta) = - \sum_i \log P \left( o^{(i)} \mid o^{(1:i-1)}, q, v \right)
\label{eq:sft}
\end{equation}

where $v$ represents the input visual information, $q$ denotes the textual query and instruction, $o^{(i)}$ represents the $i$-th token in the generated response, and $o^{(1:i-1)}$ corresponds to the preceding context tokens. This supervised learning approach establishes the foundational spatial capabilities that are subsequently refined through reinforcement learning in Stage 3.

\paragraph{Details of Cold Start.}
Before the formal GRPO training in Stage 3, we perform a cold-start~\citep{guo2025deepseek} phase to ensure that the model can more reliably generate outputs that satisfy the required format. Specifically, we adopt a rejection sampling strategy to construct chain-of-thought augmented data with composite formatting constraints. Concretely, based on the spatial reasoning tasks from \datasetname, we use the Qwen2.5-VL-7B model to generate candidate question–answer pairs with reasoning chains. The generated responses are then filtered using a reward function under two criteria: (1) the response must strictly satisfy the predefined formatting requirements, and (2) its accuracy reward must exceed a predefined threshold. The resulting rejection-sampled dataset is defined as:

\begin{equation}
    \mathcal{D_\mathrm{coldstart}}=\left\{ (v_i,q_i, {o_i}) \,\middle|\, (v_i,q_i,{o_i},y_i) \in \mathcal{D_\mathrm{candicate}}
    \land \mathcal{R}({o_i},y_i) > 1 + \lambda
    \right\}
\label{eq:cold_start}
\end{equation}

where $\mathcal{D_\mathrm{candicate}}$ denotes the set of candidate question–answer pairs generated by Qwen2.5-VL-7B on \datasetname, $v_i$ represents the visual input of the $i$-th question, $q_i$ denotes the question text, $o_i$ corresponds to the model’s response for the $i$-th question, and $\lambda$ is the accuracy reward threshold. This process yielded a total of 1,255 cold-start training samples.

\paragraph{Details of Reward Function}
Stage 3 introduces the GRPO reinforcement learning algorithm to further stimulate the model's spatial reasoning capabilities through carefully designed reward mechanisms. Our reward system includes format rewards and accuracy rewards.

The format reward ensures structured model outputs by requiring the model to place its reasoning process and final answer within \texttt{<think>} ... \texttt{</think>} and \texttt{<answer>} ... \texttt{</answer>} tags, respectively:

\begin{equation}
r_{\text{format}}(o) =
\begin{cases}
1, & \text{if } o \text{ matches format} \\
0, & \text{otherwise}
\end{cases}
\end{equation}

The accuracy reward employs differentiated evaluation strategies based on question types. For multiple-choice questions, we adopt a strict exact matching criterion:

\begin{equation}
r_{\text{mc}}(o, y) =
\mathbb{I}(o=y) \tag{3}
\end{equation}

where $o$ represents the model's prediction and $y$ denotes the ground truth label for the question.

For numerical answer questions, considering the continuous nature of numerical predictions, we design a weighted relative accuracy measure based on confidence intervals:

\begin{equation}
r_{\text{num}}(o, y) 
= \frac{1}{|\mathcal{T}|} \sum_{\tau \in \mathcal{T}}
\mathbb{I} \left( 
\frac{ \left| o - y) \right| }
     {  y }
< \tau \right)
\end{equation}

where $\mathcal{T}=[0.50, 0.55, ..., 0.95]$ represents a series of confidence thresholds.

The unified accuracy reward function is defined as:

\begin{equation}
r_\text{accuracy}(o, y)=
\begin{cases}
r_\text{mc}(o, y), & \text{if } q \in \text{MCQ} \\
r_\text{num}(o, y), & \text{if } q \in \text{NQ}
\end{cases}
\end{equation}

where $q$ represents the input question, $\text{MCQ}$ denotes the set of multiple-choice questions, $\text{NQ}$ denotes the set of numerical answer questions.

The final reward function integrates both format and accuracy dimensions:

\begin{equation}
\mathcal{R}(o, y)=r_{\mathrm{format}}(o) + r_{\mathrm{accuray}}(o,y)
\end{equation}

\section{Additional Experiments Details}

\begin{table}[H]
\caption{Detailed statistics of the SPBench-SI and SPBench-MV.}
\centering
\label{tab:bench_statics}
\resizebox{0.9\linewidth}{!}{
\begin{tabular}{lcccccc}
\toprule
\multirow{2}{*}{\textbf{Benchmark}} & \multicolumn{3}{c}{\textbf{Numerical Question}} & \multicolumn{2}{c}{\textbf{Multiple-choice Question}} & \multirow{2}{*}{\textbf{Total}} \\
\cmidrule(r){2-4} \cmidrule(l){5-6}
& Obj. Cnt. & Abs. Dist. & Obj. Size & Rel. Dist. & Rel. Dir. & \\
\midrule
SPBench-SI & - & 149 & 463 & 91 & 306 & 1,009 \\
SPBench-MV & 70 & 30 & 158 & 17 & 44 & 319 \\
\bottomrule
\end{tabular}
}
\vspace{-10pt}
\end{table}

\subsection{Details of Benchmarks}
\label{app:details_of_benchmarks}

\begin{itemize}
    \item \textbf{VSI-Bench}~\citep{yang2025thinking}: VSI-Bench is a comprehensive evaluation benchmark for assessing visual-spatial intelligence in Multimodal Large Language Models (MLLMs) through egocentric video understanding. The benchmark comprises over 5,000 question-answer pairs from 288 real-world videos sourced from ScanNet~\cite{dai2017scannet}, ScanNet++~\cite{yeshwanth2023scannet++}, and ARKitScenes~\citep{baruch2021arkitscenes}, spanning diverse environments across multiple geographic regions.
    
    \item \textbf{SPBench-SI \& SPBench-MV}: SPBench-SI and SPBench-MV are evaluation benchmarks constructed using the \datasetname{} pipeline applied to the ScanNet validation set. SPBench-SI serves as a single-image evaluation benchmark designed to assess models' spatial understanding and reasoning capabilities from individual viewpoints, encompassing four task categories: absolute distance, object size, relative distance, and relative direction, with a total of 1,009 samples. SPBench-MV constitutes a multi-view evaluation benchmark that requires models to perform joint spatial modeling across multiple viewpoints. SPBench-MV additionally incorporates object counting tasks to evaluate models' capabilities in identifying and enumerating objects within multi-view scenarios, with a total of 319 samples. Both benchmarks undergo rigorous quality control through the standard pipeline filtering strategies supplemented by manual curation to ensure data disambiguation and high-quality annotations. The detailed statistics of SPBench-SI and SPBench-MV are provided in Table~\ref{tab:bench_statics}.
    
    \item \textbf{CV-Bench}~\citep{tong2024cambrian}: CV-Bench addresses limitations of existing vision-centric benchmarks through 2,638 manually-inspected examples. The benchmark repurposes established vision datasets—ADE20k~\citep{zhou2017ade20k}, COCO~\citep{lin2014coco}, and OMNI3D~\citep{brazil2023omni3d}—to evaluate MLLMs on fundamental computer vision tasks. The evaluation encompasses 2D spatial comprehension through spatial relationships and object counting, while 3D understanding is assessed via depth ordering and relative distance estimation.
    
    \item \textbf{SPAR-Bench}~\citep{zhang2025flatland}: SPAR-Bench constitutes a comprehensive evaluation framework for systematically assessing spatial perception and reasoning capabilities in VLMs. The benchmark encompasses 20 diverse spatial understanding tasks spanning single-view, multi-view, and temporal video modalities, incorporating 7,207 manually verified question-answer pairs to ensure annotation quality and reliability.
    
    \item \textbf{ViewSpatial-Bench}~\citep{li2025viewspatial}: ViewSpatial-Bench is a comprehensive evaluation framework comprising over 5,700 question-answer pairs across 1,000+ 3D scenes from ScanNet~\citep{dai2017scannet} and MS-COCO~\citep{lin2014coco} validation datasets. This benchmark evaluates VLMs' spatial localization capabilities from both egocentric and allocentric viewpoints, addressing the critical gap in perspective-taking abilities essential for embodied interaction and multi-agent collaboration.
\end{itemize}

\subsection{Details of Baselines}
\label{app:details_of_baselines}

\begin{itemize}
    \item \textbf{GPT-4o}~\citep{hurst2024gpt}: GPT-4o (where "o" denotes "omni") is a multilingual, multimodal generative pre-trained transformer model developed by OpenAI and released in May 2024. The model demonstrates comprehensive multimodal capabilities, supporting the processing and generation of text, images, and audio across multiple languages.
    
    \item \textbf{Gemini-2.0-Flash}~\citep{team2024gemini}: Gemini 2.0 Flash is a multimodal language model designed for agent-oriented applications. The model features enhanced computational efficiency, integrated tool utilization, multimodal content generation, and a one million token context window. Building upon previous Flash architectures, it demonstrates improved performance quality while maintaining comparable inference speeds.
    
    \item \textbf{InternVL-2.5-4B/8B}~\citep{chen2024internvl}: InternVL 2.5 is an advanced MLLM building upon InternVL 2.0 with enhanced training strategies and data quality. The model demonstrates competitive performance across diverse benchmarks including reasoning, document understanding, and video comprehension, rivaling commercial models like GPT-4o and Claude-3.5-Sonnet.
    
    \item \textbf{Kimi-VL-A3B}~\citep{team2025kimi}: Kimi-VL-A3B is an efficient open-source Mixture-of-Experts (MoE)~\citep{jiang2024mixtral} VLM featuring advanced multimodal reasoning, long-context understanding, and agent capabilities while activating only 2.8B parameters in its language decoder.
    
    \item \textbf{LLaVA-Onevision-7B}~\citep{li2024llava-one}: LLaVA-OneVision-7B is an open MLLM that achieves strong performance across single-image, multi-image, and video scenarios. The model demonstrates effective transfer learning across different modalities, with particularly strong video understanding capabilities emerging through task transfer from images to videos.
    
    \item \textbf{Qwen2.5-VL-3B/7B}~\citep{bai2025qwen2}: Qwen2.5-VL is an open VLM in the Qwen series, featuring enhanced visual recognition, object localization, document parsing, and long-video comprehension. The model introduces dynamic resolution processing to handle varying image sizes and extended video durations.
    
    \item \textbf{SpaceR-7B}~\citep{ouyang2025spacer}: SpaceR is a MLLM designed for video spatial reasoning using reinforcement learning with verifiable reward. The model incorporates a map imagination mechanism to infer spatial layouts during reasoning and achieves competitive performance, significantly surpassing GPT-4o on VSI-Bench.
    
    \item \textbf{VILASR-7B}~\citep{wu2025reinforcing}: VILASR introduces a "drawing to reason in space" paradigm that enables VLMs to perform spatial reasoning through elementary drawing operations like annotating bounding boxes and drawing auxiliary lines. The model uses three-stage training: cold-start with synthetic data, reflective rejection sampling, and reinforcement learning for reward optimization. VILASR outperforms existing methods across spatial reasoning benchmarks.
    
    \item \textbf{Video-R1}~\citep{feng2025video}: Video-R1 applies the R1 reasoning paradigm to video understanding in MLLMs, following DeepSeek-R1's~\citep{guo2025deepseek} reinforcement learning approach. The model employs the T-GRPO algorithm to enhance temporal information utilization and trains on both image and video reasoning data. Video-R1-7B achieves strong performance on video reasoning benchmarks, surpassing GPT-4o on VSI-Bench and demonstrating robust capabilities across general video tasks
    
    \item \textbf{Spaital-MLLM-4B}~\citep{wu2025spatial}: Spatial-MLLM is a novel framework for visual-based spatial reasoning. The model employs a dual-encoder architecture combining a pretrained 2D visual encoder for semantic features with a spatial encoder initialized from a visual geometry foundation model for 3D structure features. Extensive experiments demonstrate state-of-the-art performance across various visual-based spatial understanding and reasoning tasks.
\end{itemize}

\begin{table}[H]
\centering
\small
\caption{\textbf{Evaluation results on VSI-Bench.} For each metric, \textbf{bold} numbers indicate the best performance, while \underline{underlined} numbers represent the second-best performance.}
\label{tab:vsibench}
\resizebox{\textwidth}{!}{
\begin{tabular}{lccccccccc}
\toprule
\multirow{2}{*}{\textbf{Model}} & \multicolumn{4}{c}{\textbf{Numerical Question}} & \multicolumn{4}{c}{\textbf{Multiple-choice Question}} & \multirow{2}{*}{\textbf{Avg.}} \\
\cmidrule(lr){2-5} \cmidrule(lr){6-9}
& Obj. Cnt & Abs. Dist. & Obj. Size & Room Size & Rel. Dist. & Rel. Dir. & Route Plan. & Appr. Order & \\
\midrule
\rowcolor{gray!10} 
\multicolumn{10}{l}{\textit{\textbf{Proprietary Models}}} \\
GPT-4o & 46.2 & 5.3 & 43.8 & 38.2 & 37.0 & 41.3 & 31.5 & 28.5 & 34.0 \\
Gemini-2.0-Flash & 56.2 & 30.9 & \textbf{66.7} & 31.8 & \textbf{51.3} & \underline{46.3} & 24.5 & \textbf{55.1} & 45.4 \\
\midrule
\rowcolor{gray!10} 
\multicolumn{10}{l}{\textit{\textbf{Open-Source Models}}} \\
InternVL-2.5-4B & 45.0 & 15.5 & 37.5 & 24.6 & 37.2 & 41.5 & 31.4 & 26.2 & 32.6 \\
InternVL-2.5-8B & 50.6 & 31.3 & 40.2 & 39.3 & 45.1 & 41.4 & 29.4 & 43.9 & 40.2 \\
Kimi-VL-A3B  & 41.3 & 30.4 & 42.1 & 13.2 & 26.3 & 32.6 & 32.0 & 11.2 & 28.7 \\
LLaVA-OneVision-7B & 46.1 & 26.2 & 36.3 & 29.5 & 30.8 & 37.2 & \underline{35.1} & 21.8 & 33.1 \\
\midrule
\rowcolor{gray!10} 
\multicolumn{10}{l}{\textit{\textbf{Qwen2.5-VL-7B Based Spatial Models}}} \\
Qwen2.5-VL-7B & 43.5 & 15.1 & 48.5 & \underline{41.1} & 36.3 & 40.1 & 28.4 & 33.7 & 35.8 \\
SpaceR-7B & 63.2 & 30.0 & 60.3 & 37.6 & 39.7 & 45.6 & 31.4 & 48.2 & 44.5 \\
VILASR-7B & 63.5 & \underline{34.4} & 60.6 & 30.9 & \underline{48.9} & 45.2 & 30.4 & 49.2 & 45.5 \\
Video-R1 & 34.0 & 23.0 & 41.6 & 36.7 & 36.8 & 34.7 & 31.4 & 28.8 & 33.4 \\
\midrule
\rowcolor{gray!10} 
\multicolumn{10}{l}{\textit{\textbf{Qwen2.5-VL-3B Based Spatial Models}}} \\
Qwen2.5-VL-3B & 32.9 & 22.1 & 17.3 & 31.5 & 32.8 & 44.2 & 26.3 & 28.5 & 29.4 \\
Spatial-MLLM-4B & \textbf{65.6} & \textbf{35.5} & \underline{64.2} & 40.6 & 41.3 & \textbf{47.9} & 34.0 & \underline{49.2} & \textbf{47.3} \\
\methodname-3B & \underline{63.5} & 34.3 & 61.7 & \textbf{43.9} & 45.4 & 44.8 & \textbf{35.6} & 36.4 & \underline{45.7} \\
\textcolor{ForestGreen}{\textit{\textbf{Improvement}}} & \textcolor{ForestGreen}{+30.6} & \textcolor{ForestGreen}{+12.2} & \textcolor{ForestGreen}{+44.4} & \textcolor{ForestGreen}{+12.4} & \textcolor{ForestGreen}{+12.6} & \textcolor{ForestGreen}{+0.6} & \textcolor{ForestGreen}{+9.3} & \textcolor{ForestGreen}{+7.9} & \textcolor{ForestGreen}{+16.3} \\
\bottomrule
\end{tabular}
}
\vspace{-10pt}
\end{table}

\begin{table}[H]
\centering
\small
\caption{\textbf{Evaluation results on SPBench-SI.} For each metric, \textbf{bold} numbers indicate the best performance, while \underline{underlined} numbers represent the second-best performance.}
\label{tab:spbench-si}
\resizebox{0.65\textwidth}{!}{
\begin{tabular}{lccccc}
\toprule
\multirow{2}{*}{\textbf{Model}} & \multicolumn{2}{c}{\textbf{Numerical Question}} & \multicolumn{2}{c}{\textbf{Multiple-choice Question}} & \multirow{2}{*}{\textbf{Avg.}} \\
\cmidrule(lr){2-3} \cmidrule(lr){4-5}
& Abs. Dist. & Obj. Size & Rel. Dist. & Rel. Dir. & \\
\midrule
\rowcolor{gray!10} 
\multicolumn{6}{l}{\textit{\textbf{Proprietary Models}}} \\
GPT-4o & 19.7 & 29 & 81.3 & 39.2 & 42.4 \\
Gemini-2.0-Flash & \underline{33.1} & \underline{64.9} & 81.3 & 39.5 & \underline{54.7} \\
\midrule
\rowcolor{gray!10} 
\multicolumn{6}{l}{\textit{\textbf{Open-Source Models}}} \\
InternVL-2.5-4B & 27.3 & 36.2 & 73.6 & 33.0 & 42.5 \\
InternVL-2.5-8B & 15.6 & 40.8 & 76.9 & 35.6 & 42.3 \\
Kimi-VL-A3B  & 11.3 & 40.2 & 62.6 & 27.1 & 35.3 \\
LLaVA-OneVision-7B & 23.6 & 27.2 & 54.9 & 27.1 & 33.2 \\
\midrule
\rowcolor{gray!10} 
\multicolumn{6}{l}{\textit{\textbf{Qwen2.5-VL-7B Based Spatial Models}}} \\
Qwen2.5-VL-7B & 27.7 & 45.0 & \textbf{83.5} & 37.6 & 48.4 \\
SpaceR-7B & 8.4 & 62.9 & 80.2 & 42.8 & 48.6 \\
VILASR-7B & 10.3 & 63.0 & 81.3 & \underline{46.1} & 50.2 \\
Video-R1 & 5.1 & 50.3 & \underline{82.4} & 41.5 & 44.8 \\
\midrule
\rowcolor{gray!10} 
\multicolumn{6}{l}{\textit{\textbf{Qwen2.5-VL-3B Based Spatial Models}}} \\
Qwen2.5-VL-3B & 30.9 & 17.8 & 75.8 & 36.6 & 40.3 \\
Spatial-MLLM-4B & 16.4 & 59.7 & 69.2 & 29.4 & 43.7 \\
\methodname-3B & \textbf{45.5} & \textbf{71.7} & 81.3 & \textbf{82.4} & \textbf{70.2} \\
\textcolor{ForestGreen}{\textit{\textbf{Improvement}}} & \textcolor{ForestGreen}{+14.6} & \textcolor{ForestGreen}{+53.9} & \textcolor{ForestGreen}{+5.5} & \textcolor{ForestGreen}{+45.8} & \textcolor{ForestGreen}{+29.9} \\
\bottomrule
\end{tabular}
}
\vspace{-10pt}
\end{table}

\begin{table}[H]
\centering
\small
\caption{\textbf{Evaluation results on SPBench-MV.} For each metric, \textbf{bold} numbers indicate the best performance, while \underline{underlined} numbers represent the second-best performance.}
\label{tab:spbench-mv}
\resizebox{0.73\textwidth}{!}{
\begin{tabular}{lcccccc}
\toprule
\multirow{2}{*}{\textbf{Model}} & \multicolumn{3}{c}{\textbf{Numerical Question}} & \multicolumn{2}{c}{\textbf{Multiple-choice Question}} & \multirow{2}{*}{\textbf{Avg.}} \\
\cmidrule(lr){2-4} \cmidrule(lr){5-6}
& Obj. Cnt & Abs. Dist. & Obj. Size & Rel. Dist. & Rel. Dir. & \\
\midrule
\rowcolor{gray!10} 
\multicolumn{7}{l}{\textit{\textbf{Proprietary Models}}} \\
GPT-4o & 66.3 & 12.0 & 43.8 & 82.4 & 36.4 & 48.2 \\
Gemini-2.0-Flash & 49.9 & \textbf{40.7} & 65.1 & 76.5 & 25.0 & 51.4 \\
\midrule
\rowcolor{gray!10} 
\multicolumn{7}{l}{\textit{\textbf{Open-Source Models}}} \\
InternVL-2.5-4B & 65.1 & 24.0 & 23.9 & 82.4 & 20.5 & 43.2 \\
InternVL-2.5-8B & 50.0 & 25.0 & 37.0 & 88.2 & 6.8 & 41.4 \\
Kimi-VL-A3B & 13.7 & 23.3 & 33.0 & 76.5 & \underline{38.6} & 37.0 \\
LLaVA-OneVision-7B & 21.1 & 21.3 & 19.2 & 76.5 & 22.7 & 32.2 \\
\midrule
\rowcolor{gray!10} 
\multicolumn{7}{l}{\textit{\textbf{Qwen2.5-VL-7B Based Spatial Models}}} \\
Qwen2.5-VL-7B & 46.1 & 11.0 & 35.4 & 88.2 & 11.4 & 37.3 \\
SpaceR-7B & \underline{90.1} & 33.7 & 65.1 & 82.4 & 25.0 & 59.4 \\
VILASR-7B & 65.3 & 34.7 & \underline{68.7} & 88.2 & 29.5 & 61.8 \\
Video-R1 & 33.9 & 18.0 & 45.6 & 76.5 & 29.5 & 40.7 \\
\midrule
\rowcolor{gray!10} 
\multicolumn{7}{l}{\textit{\textbf{Qwen2.5-VL-3B Based Spatial Models}}} \\
Qwen2.5-VL-3B & 36.7 & 14.7 & 14.9 & 88.2 & 18.2 & 36.6 \\
Spatial-MLLM-4B & 88.9 & 31.0 & 71.2 & \underline{88.2} & 29.5 & \underline{61.8} \\
\methodname-3B & \textbf{94.9} & \underline{34.7} & \textbf{76.4} & \textbf{100} & \textbf{50.0} & \textbf{71.2} \\
\textcolor{ForestGreen}{\textit{\textbf{Improvement}}} & \textcolor{ForestGreen}{+58.2} & \textcolor{ForestGreen}{+20.0} & \textcolor{ForestGreen}{+61.5} & \textcolor{ForestGreen}{+11.8} & \textcolor{ForestGreen}{+31.8}  & \textcolor{ForestGreen}{+34.6}\\
\bottomrule
\end{tabular}
}
\vspace{-10pt}
\end{table}

\subsection{Details of In-domain Benchmarks Results}
\label{app:in-domain_details}

We present the detailed evaluation results of VSI-Bench in Table \ref{tab:vsibench}. Our proposed \methodname{} achieves an overall accuracy of 45.7\%, surpassing all compared models except Spatial-MLLM, including those with 2–3 times larger parameter sizes. On average, \methodname{} improves performance by 16.3\% and demonstrates consistent gains across all sub-tasks of VSI-Bench. Notably, while Spatial-MLLM leverages an additional 3D encoder, our \methodname{} relies solely on the vision encoder of Qwen2.5-VL-3B.

Furthermore, we report the detailed results of SPBench-SI and SPBench-MV in Tables \ref{tab:spbench-si} and \ref{tab:spbench-mv}, respectively. \methodname{} attains 70.2\% and 71.2\% accuracy on these two benchmarks, corresponding to relative improvements of 29.9\% and 34.6\% over the base model Qwen2.5-VL-3B, and consistently outperforms all compared baselines.

\begin{table}[H]
\centering
\small
\caption{\textbf{Evaluation results on CV-Bench.} For each metric, \textbf{bold} numbers indicate the best performance, while \underline{underlined} numbers represent the second-best performance.}
\label{tab:cv-bench}
\resizebox{0.62\textwidth}{!}{
\begin{tabular}{lccccc}
\toprule
\multirow{2}{*}{\textbf{Model}} & \multicolumn{3}{c}{\textbf{2D}} & \multicolumn{1}{c}{\textbf{3D}} & \multirow{2}{*}{\textbf{Overall}} \\
\cmidrule(lr){2-4}
& ADE20K & COCO & Avg. & Omni3D & \\
\midrule
GPT-4o & 65.1 & 73.8 & 69.4 & \underline{81.3} & \underline{75.4} \\
InternVL-2.5-4B & \underline{68.6} & \underline{78.5} & \underline{73.5} & 75.1 & 74.4 \\
Kimi-VL-A3B & 41.9 & 42.4 & 41.9 & 54.7 & 48.3 \\
LLaVA-OneVision-7B & 50.6 & 55.8 & 53.2 & 63.5 & 58.3 \\
Qwen2.5-VL-7B & \textbf{69.5} & \textbf{80.5} & \textbf{75.0} & \textbf{83.1} & \textbf{79.0} \\
\midrule
\rowcolor{gray!10}
Qwen2.5-VL-3B & 63.2 & 75.0 & 69.1 & 72.2 & 70.6 \\
\rowcolor{gray!10}
\methodname-3B & 67.1 & 77.6 & 72.4 & 74.9 & 73.7 \\
\rowcolor{gray!10}
\textcolor{ForestGreen}{\textit{\textbf{Improvement}}} & \textcolor{ForestGreen}{+3.9} & \textcolor{ForestGreen}{+2.6} & \textcolor{ForestGreen}{+3.3} & \textcolor{ForestGreen}{+2.7} & \textcolor{ForestGreen}{+3.1} \\
\bottomrule
\end{tabular}
}
\vspace{-10pt}
\end{table}

\begin{table}[H]
\centering
\small
\caption{\textbf{Evaluation results on ViewSpatial-Bench.} For each metric, \textbf{bold} numbers indicate the best performance, while \underline{underlined} numbers represent the second-best performance.}
\label{tab:vs-bench}
\resizebox{0.9\textwidth}{!}{
\begin{tabular}{lcccccccc}
\toprule
\multirow{2}{*}{\textbf{Model}} & \multicolumn{3}{c}{\textbf{Camera Perspective}} & \multicolumn{4}{c}{\textbf{Person Perspective}} & \multirow{2}{*}{\textbf{Overall}} \\
\cmidrule(lr){2-4} \cmidrule(lr){5-8}
& Rel. Dir. & Obj. Ori. & Avg. & Obj. Ori. & Rel. Dir. & Sce. Sim. & Avg. \\
\midrule
GPT-4o & 41.5 & 19.6 & 33.7 & 41.2 & 32.8 & 21.9 & 31.5 & 32.6 \\
InternVL-2.5-4B & 37.1 & \underline{31.8} & \underline{40.8} & 43.6 & \underline{37.1} & 26.1 & 35.1 & 37.9 \\
Kimi-VL-A3B & 26.9 & 22.1 & 25.1 & \underline{63.1} & \textbf{43.9} & 20.3 & 41.5 & 33.6 \\
LLaVA-OneVision-7B & 29.8 & 26.1 & 28.5 & 22.4 & 31.0 & 26.9 & 26.5 & 27.5 \\
Qwen2.5-VL-7B & \underline{47.8} & 30.9 & \textbf{41.8} & 41.6 & 35.4 & \underline{26.9} & \underline{39.8} & \underline{37.9} \\
\midrule
\rowcolor{gray!10}
Qwen2.5-VL-3B & 43.5 & \textbf{32.5} & 39.5 & 40.0 & 29.9 & 26.3 & 32.0 & 35.6 \\
\rowcolor{gray!10}
\methodname-3B & \textbf{48.3} & 24.1 & 39.6 & \textbf{71.1} & 34.4 & \textbf{38.9} & \textbf{48.5} & \textbf{44.2} \\
\rowcolor{gray!10}
\textcolor{ForestGreen}{\textit{\textbf{Improvement}}} & \textcolor{ForestGreen}{+4.8} & \textcolor{DarkRed}{-8.4} & \textcolor{ForestGreen}{+0.1} & \textcolor{ForestGreen}{+31.1} & \textcolor{ForestGreen}{+4.5} & \textcolor{ForestGreen}{+12.6} & \textcolor{ForestGreen}{+16.5} & \textcolor{ForestGreen}{+8.6}\\
\bottomrule
\end{tabular}
}
\vspace{-10pt}
\end{table}

\subsection{Details of Out-of-domain Benchmarks Results}
\label{app:out-of-domain_details}

We present the detailed evaluation results on CV-Bench in Table \ref{tab:cv-bench}. Our proposed \methodname{} achieves an overall performance of 73.7\%, 5.3\% below the best-performing baseline model. Nevertheless, it surpasses the base model Qwen2.5-VL-3B by 3.1\%, demonstrating the robustness of our training framework in enhancing model performance. The detailed results on ViewSpatial-Bench are provided in Table~\ref{tab:vs-bench}, where \methodname{} achieves an overall accuracy of 44.2\%, outperforming all compared models and surpassing the base model by 8.6\%. The detailed evaluation results on the out-of-domain benchmark SPAR-Bench are provided in Table \ref{tab:out-of-domain} in the main text.

\subsection{Details of Semantic Entropy.}
\label{app:entropy_details}

To quantify response diversity for uncertainty analysis, we introduce semantic entropy as a clustering-based metric. For each question $q$, we sample responses $\{o_1,o_2,...,o_G\}$ at temperature 0.9 (8 samples per question) and partition them into semantic clusters $\mathcal{C} = \{C_1, C_2, ..., C_K\}$ based on accuracy rewards, where $C_k = \{o_i : \mathcal{R}(o_i) = r_k\}$ for distinct reward values $\mathcal{R} = \{r_1, r_2, ..., r_K\}$. Semantic entropy is then computed as:
\begin{equation}
\text{SE}(q) = - \sum_{i=1}^K p_i \log p_i, \quad \text{where } p_i = \frac{|C_i|}{N}
\end{equation}
This measure captures the distributional diversity of semantically distinct response clusters, providing a principled approach to quantify model uncertainty beyond surface-level textual variations.

\subsection{Details of Attention Analysis.}
\label{app:attention_details}

For visual attention analysis, we use two metrics, Visual Attention IoU, used to measures attention concentration within object bounding boxes, defined as:
\begin{equation}
\text{IoU}_{\text{att}} = \frac{\sum_{i \in B_{\text{obj}}} \hat{a}_i}{\sum_{j=1}^{N} \hat{a}_j}
\end{equation}
where $B_{\text{obj}}$ represents visual tokens within the target object's bounding box and $\hat{a}_i$ denotes min-max normalized attention weights. Visual Attention Entropy measures attention concentration:
\begin{equation}
H_{\text{att}} = -\sum_{i=1}^{N} p_i \log p_i
\end{equation}
where $p_i$ represents the probability distribution from normalized attention weights.

\begin{figure}[t]
	\centering
	\includegraphics[width=1\linewidth]{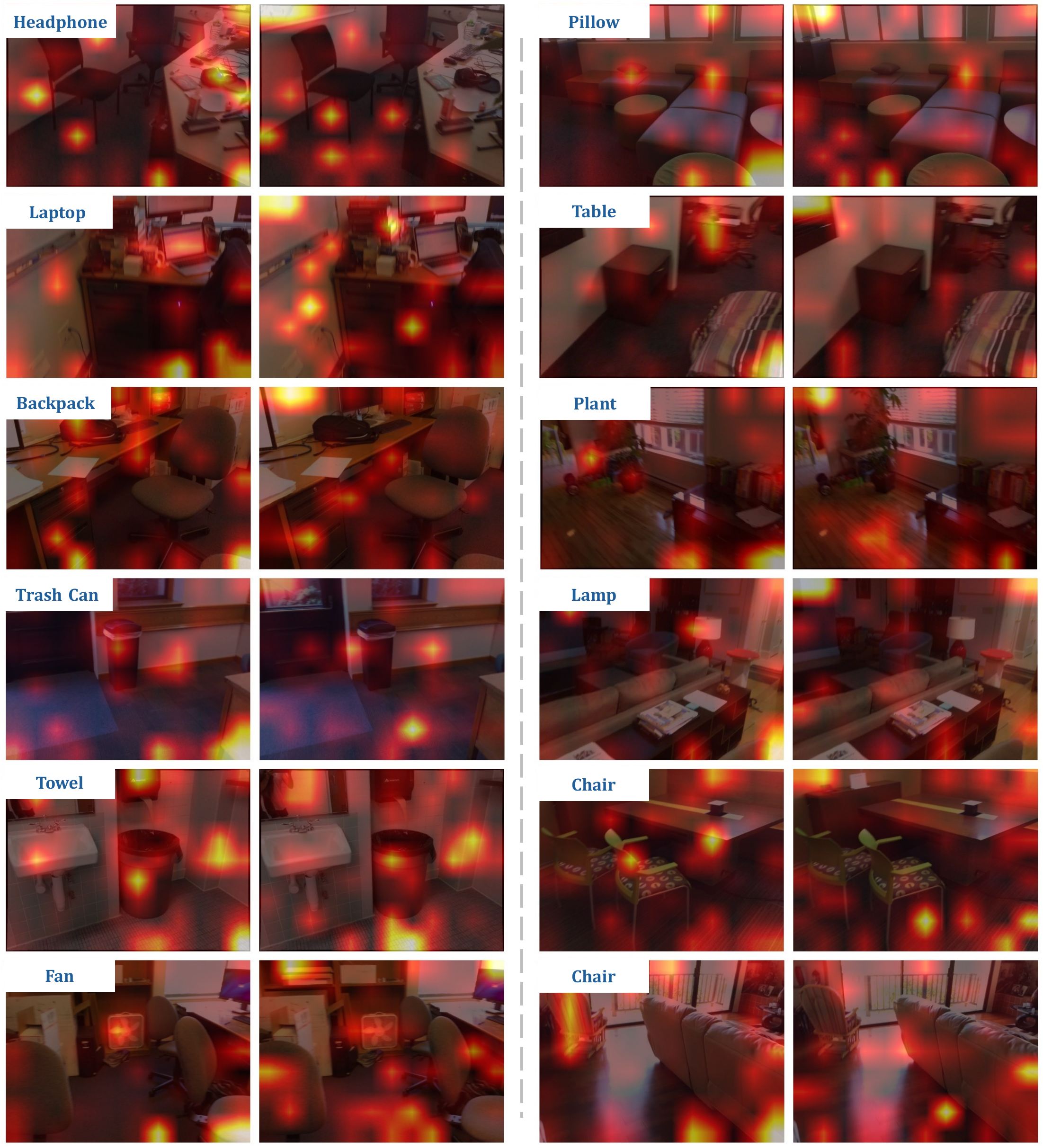}
	\caption{\textbf{Additional examples of attention distribution comparison.} For each example, the left panel shows the attention distribution of \methodname, while the right panel shows that of Qwen2.5-VL-3B.}
	\label{fig:attention_cases}
    \vspace{-5pt}
\end{figure}

In Figure~\ref{fig:attention_cases}, we present additional comparisons of attention distributions between \methodname{} and its base model Qwen2.5-VL-3B on the object size estimation task. The results demonstrate that, compared with the base model, \methodname{} exhibits a more focused allocation of attention on task-relevant objects. This indicates that our training framework effectively guides the internal attention distribution of the model, thereby enhancing its inherent perceptual ability and supporting more reliable performance in spatial reasoning tasks.

\begin{figure}[t]
	\centering
	\includegraphics[width=1\linewidth]{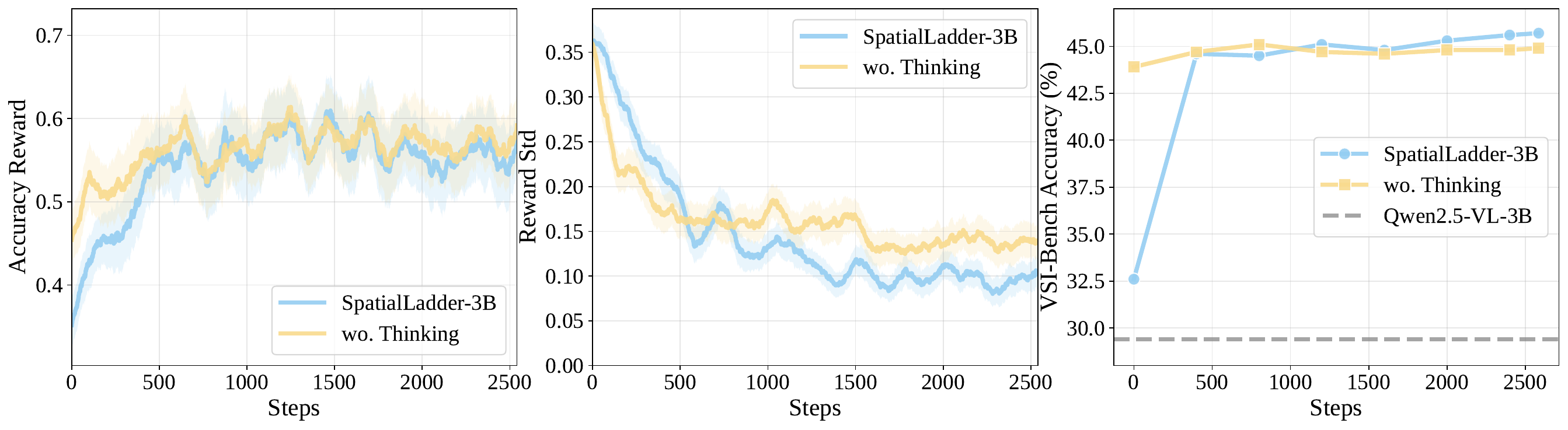}
	\caption{\textbf{Impact of thinking during training.} Left: accuracy rewards; Middle: reward standard deviation; Right: VSI-Bench performance comparison.}
	\label{fig:think_dynamics}
    \vspace{-10pt}
\end{figure}

\section{Additional Experiments}

\subsection{Chain-of-thought Training Dynamics}
\label{app:cot_dynamics}

The analysis of chain-of-thought reasoning dynamics, as illustrated in Figure \ref{fig:think_dynamics}, provides additional insights into the importance of explicit reasoning processes in spatial understanding tasks. While both the full model with chain-of-thought and its variant without reasoning components achieve comparable performance in terms of accuracy reward curves during later training stages, significant differences emerge in training stability and convergence patterns.The chain-of-thought enabled model demonstrates superior training stability, characterized by faster reduction in reward standard deviation and smoother convergence behavior. More critically, the actual performance trajectories reveal distinct learning dynamics: the model without chain-of-thought reasoning reaches an early performance plateau and exhibits limited improvement thereafter, whereas the chain-of-thought variant maintains continuous performance enhancement throughout training, ultimately achieving superior final accuracy on evaluation benchmarks.

\begin{table}[t]
\caption{\textbf{Performance comparison across spatial reasoning datasets.} SI, MV, and VID denote single-image, multi-view, and video modalities, respectively.}
\centering
\label{tab:dataset_comparison}
\resizebox{0.8\textwidth}{!}{
\begin{tabular}{lccc}
\toprule
\textbf{Dataset} & \textbf{Modality} & \textbf{Size} & \textbf{VSI-Bench} \\ \midrule
SpaceR-151$k$~\citep{ouyang2025spacer} & VID & 151,310 & 35.1 \\
Spatial-MLLM-120$k$\citep{ouyang2025spacer} & MV+VID & $\approx$120,000 & 40.0 \\
\datasetname{} & \textbf{SI+MV+VID} & \textbf{26,610} & \textbf{43.9} \\ \bottomrule
\end{tabular}
}
\vspace{-10pt}
\end{table}

\subsection{Comparison with Other Spatial Dataset}
\label{app:spatial_data_comparison}

Table~\ref{tab:dataset_comparison} demonstrates the effectiveness of our dataset design through comparative analysis with existing spatial reasoning datasets. All models are trained using supervised fine-tuning on Qwen2.5-VL-3B as the base model to ensure fair comparison. Despite utilizing significantly fewer training samples (26,610 vs. 151,310 and $\approx$120,000), \datasetname{} achieves superior performance on VSI-Bench, reaching 43.9\% accuracy compared to 35.1\% for SpaceR-151k and 40.0\% for Spatial-MLLM-120k. This performance gain is attributed to our comprehensive approach that integrates object localization tasks and spatial reasoning tasks across single-image, multi-view, and video modalities within a unified framework, contrasting with previous datasets that focus on individual modalities or limited combinations. The results validate that strategic dataset curation and progressive training can achieve better spatial reasoning capabilities with substantially reduced data requirements, highlighting the importance of data quality and training methodology over sheer dataset scale.

\begin{figure}[H]
  \centering
  \makebox[\textwidth][c]{%
      \begin{minipage}{0.63\textwidth}
        \centering
        \includegraphics[height=5.4cm]{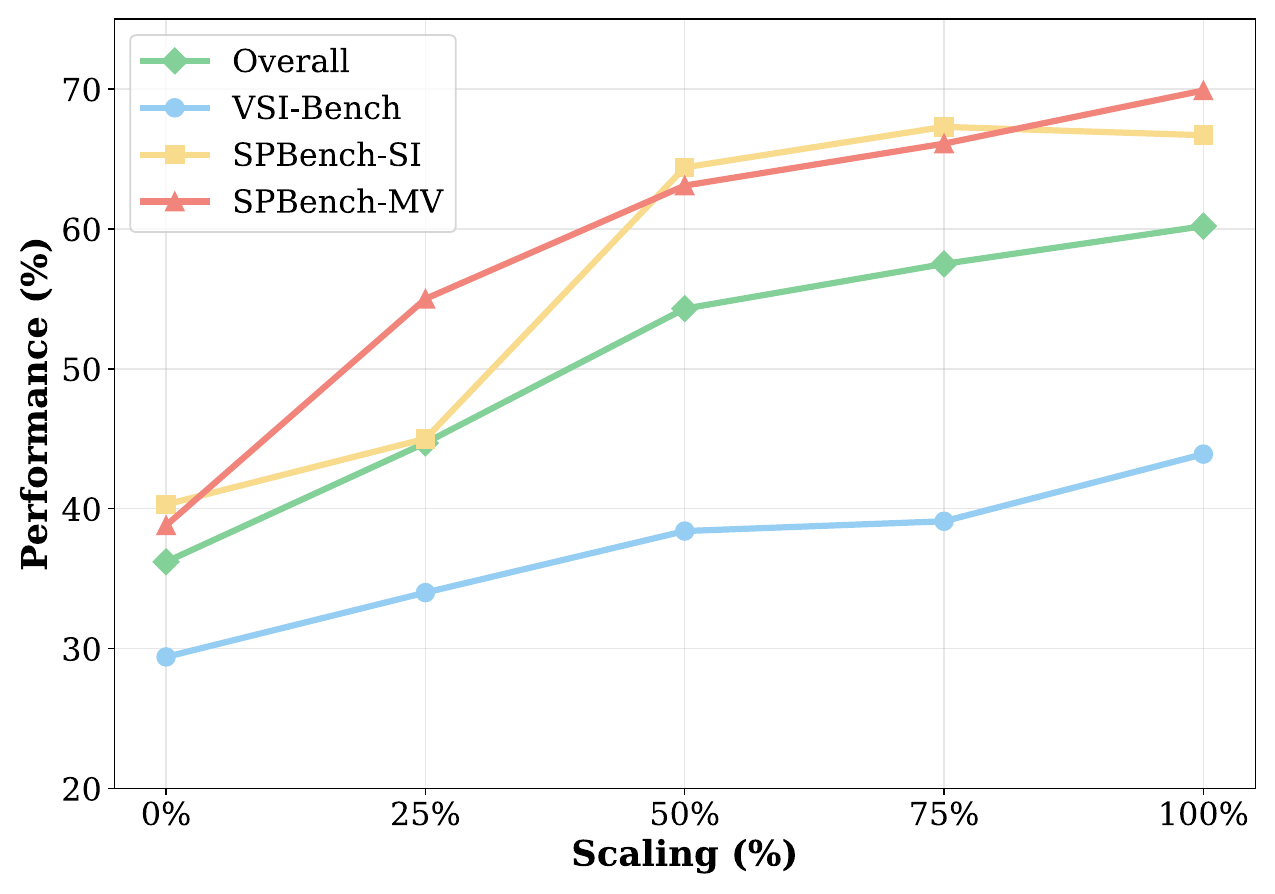}
        \caption{Dataset scaling analysis across spatial reasoning benchmarks.}
        \label{fig:data_scaling}
      \end{minipage}
      \hfill
      \begin{minipage}{0.33\textwidth}
        \centering
        \includegraphics[height=5.4cm]{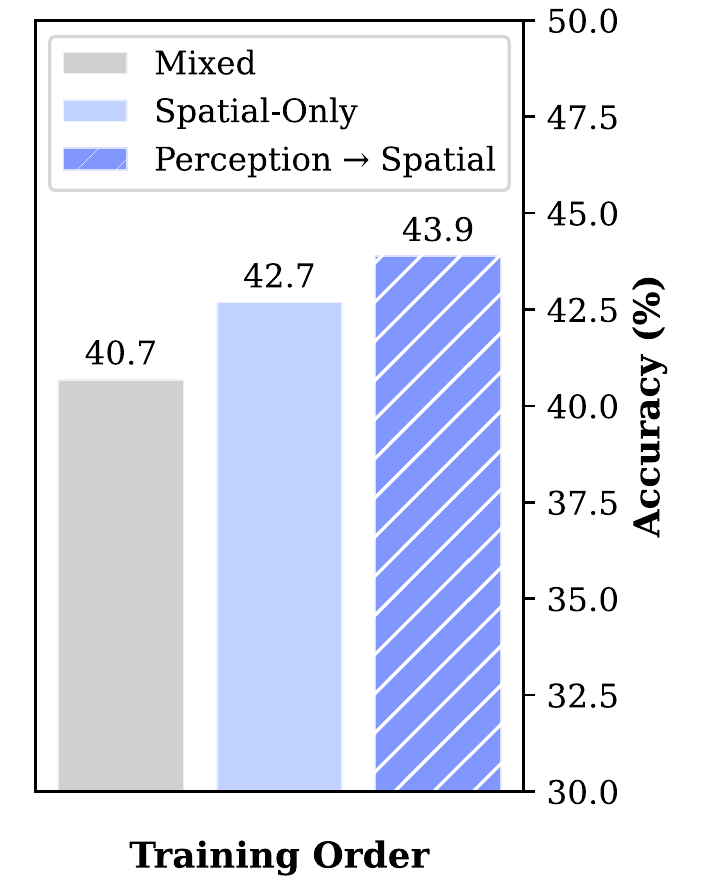}
        \caption{VSI-Bench accuracy across different training order.}
        \label{fig:progressive}
      \end{minipage}
  }
  \vspace{-10pt}
\end{figure}

\subsection{Dataset Scaling Analysis} 
\label{app:dataset_scaling}

Figure~\ref{fig:data_scaling} demonstrates the consistent scaling potential of our dataset across spatial reasoning benchmarks using Qwen2.5-VL-3B as the base model for supervised fine-tuning. Overall performance increases steadily from 36.2\% to 60.2\%, while VSI-Bench improves from 29.4\% to 43.9\% as dataset scaling progresses from 0\% to 100\%. The sustained upward trajectories without saturation at full scale indicate substantial room for further improvement through continued dataset expansion. These scaling patterns validate the effectiveness of our dataset design at larger scales and highlight the potential for achieving even stronger spatial reasoning capabilities through strategic dataset augmentation.

\subsection{Impact of Progressive Training Sequence}
\label{app:progressive}

To validate our training paradigm, we conduct ablation studies using Qwen2.5-VL-3B as the base model with supervised fine-tuning protocols. Figure~\ref{fig:progressive} demonstrates the critical importance of training order in developing spatial reasoning capabilities. Our progressive perception-to-spatial training paradigm achieves 43.9\% accuracy on VSI-Bench, outperforming both spatial-only training (42.7\%) and mixed training approaches (40.7\%). The sequential approach of first establishing perceptual foundations through localization tasks, followed by spatial training, yields 1.2\% improvement over direct spatial training and 3.2\% gain over simultaneous mixed training. These results validate our hypothesis that systematic progression from basic perception to complex reasoning creates more robust spatial understanding than alternative training strategies. The performance hierarchy clearly indicates that structured skill development through sequential training stages is essential for optimal spatial reasoning capabilities.

\section{Limitations and Future Work}
\label{app:limitations}

Our work presents several limitations. Due to computational resource constraints, our experiments are conducted exclusively on 3B-parameter models, leaving the scalability to larger models unexplored. Additionally, our \datasetname{} dataset has substantial room for scaling, with only 26,610 samples that may be insufficient for capturing the full complexity of spatial reasoning scenarios. The dataset's reliance primarily on ScanNet scenes also introduces domain bias toward indoor environments, limiting generalization to diverse real-world scenarios. Furthermore, our three-stage progressive training framework follows a fixed sequential structure that may not be optimal for all spatial reasoning tasks, lacking the flexibility to adapt to task-specific requirements.

These limitations suggest promising directions for future work. Scaling our progressive training approach to larger models (7B, 13B, and beyond) could reveal additional performance gains and better understand the scalability of hierarchical spatial learning. Expanding the dataset both in scale and diversity—incorporating larger sample sizes, outdoor landscapes, urban environments, and domain-specific imagery—would likely yield better performance and enhance robustness across varied scenarios. Additionally, developing adaptive training frameworks that dynamically adjust learning sequences based on task characteristic or model performance could improve efficiency. Finally, validating \methodname{} in real-world applications such as robotics navigation and autonomous driving would provide valuable insights into practical deployment and identify areas for further improvement.

\section*{LLM Usage}
\label{app:use_of_llms}

In this section, we clarify the role of large language models (LLMs) in preparing this work. We acknowledge that LLMs were employed exclusively for writing assistance and linguistic refinement in the preparation of this manuscript. These tools were utilized to enhance the clarity, grammatical accuracy, and academic style of the text while preserving all original research contributions, methodological approaches, and scientific insights developed by the authors. The language models served solely as writing aids to improve sentence structure, enhance readability of technical content, refine academic terminology, and ensure consistency in writing style throughout the manuscript.

It is important to emphasize that LLMs were not employed for research ideation, conceptual development, literature review, citation discovery, data analysis, experimental design, or generation of research hypotheses and conclusions. All research ideas, experimental work, data analysis, and scientific conclusions presented in this paper originate entirely from the authors' independent intellectual work. The use of LLMs was limited to linguistic enhancement and does not constitute contribution at the level of authorship. The authors take full responsibility for all content, including any text that was refined with LLM assistance, ensuring that the core intellectual contributions and scientific merit of this work remain wholly attributable to the listed authors.

%% file: main.bbl
\begin{thebibliography}{52}
\providecommand{\natexlab}[1]{#1}
\providecommand{\url}[1]{\texttt{#1}}
\expandafter\ifx\csname urlstyle\endcsname\relax
  \providecommand{\doi}[1]{doi: #1}\else
  \providecommand{\doi}{doi: \begingroup \urlstyle{rm}\Url}\fi

\bibitem[Bai et~al.(2025)Bai, Chen, Liu, Wang, Ge, Song, Dang, Wang, Wang, Tang, et~al.]{bai2025qwen2}
Shuai Bai, Keqin Chen, Xuejing Liu, Jialin Wang, Wenbin Ge, Sibo Song, Kai Dang, Peng Wang, Shijie Wang, Jun Tang, et~al.
\newblock Qwen2. 5-vl technical report.
\newblock \emph{arXiv preprint arXiv:2502.13923}, 2025.

\bibitem[Baruch et~al.(2021)Baruch, Chen, Dehghan, Dimry, Feigin, Fu, Gebauer, Joffe, Kurz, Schwartz, et~al.]{baruch2021arkitscenes}
Gilad Baruch, Zhuoyuan Chen, Afshin Dehghan, Tal Dimry, Yuri Feigin, Peter Fu, Thomas Gebauer, Brandon Joffe, Daniel Kurz, Arik Schwartz, et~al.
\newblock Arkitscenes: A diverse real-world dataset for 3d indoor scene understanding using mobile rgb-d data.
\newblock \emph{arXiv preprint arXiv:2111.08897}, 2021.

\bibitem[Brazil et~al.(2023)Brazil, Kumar, Straub, Ravi, Johnson, and Gkioxari]{brazil2023omni3d}
Garrick Brazil, Abhinav Kumar, Julian Straub, Nikhila Ravi, Justin Johnson, and Georgia Gkioxari.
\newblock Omni3d: A large benchmark and model for 3d object detection in the wild.
\newblock In \emph{Proceedings of the IEEE/CVF conference on computer vision and pattern recognition}, pp.\  13154--13164, 2023.

\bibitem[Chandrasegaran et~al.(2024)Chandrasegaran, Gupta, Hadzic, Kota, He, Eyzaguirre, Durante, Li, Wu, and Fei-Fei]{chandrasegaran2024hourvideo}
Keshigeyan Chandrasegaran, Agrim Gupta, Lea~M Hadzic, Taran Kota, Jimming He, Crist{\'o}bal Eyzaguirre, Zane Durante, Manling Li, Jiajun Wu, and Li~Fei-Fei.
\newblock Hourvideo: 1-hour video-language understanding.
\newblock \emph{Advances in Neural Information Processing Systems}, 37:\penalty0 53168--53197, 2024.

\bibitem[Chen et~al.(2025)Chen, Li, Xi, Zeng, and Wang]{chen2025perceptionreasoningtwostagereinforcement}
Yan Chen, Long Li, Teng Xi, Long Zeng, and Jingdong Wang.
\newblock Perception before reasoning: Two-stage reinforcement learning for visual reasoning in vision-language models, 2025.
\newblock URL \url{https://arxiv.org/abs/2509.13031}.

\bibitem[Chen et~al.(2024)Chen, Wang, Cao, Liu, Gao, Cui, Zhu, Ye, Tian, Liu, et~al.]{chen2024internvl}
Zhe Chen, Weiyun Wang, Yue Cao, Yangzhou Liu, Zhangwei Gao, Erfei Cui, Jinguo Zhu, Shenglong Ye, Hao Tian, Zhaoyang Liu, et~al.
\newblock Expanding performance boundaries of open-source multimodal models with model, data, and test-time scaling.
\newblock \emph{arXiv preprint arXiv:2412.05271}, 2024.

\bibitem[Dai et~al.(2017)Dai, Chang, Savva, Halber, Funkhouser, and Nie{\ss}ner]{dai2017scannet}
Angela Dai, Angel~X Chang, Manolis Savva, Maciej Halber, Thomas Funkhouser, and Matthias Nie{\ss}ner.
\newblock Scannet: Richly-annotated 3d reconstructions of indoor scenes.
\newblock In \emph{Proceedings of the IEEE conference on computer vision and pattern recognition}, pp.\  5828--5839, 2017.

\bibitem[Fan et~al.(2025)Fan, He, Yang, Zheng, Kuo, Zheng, Narayanaraju, Guan, and Wang]{fan2025grit}
Yue Fan, Xuehai He, Diji Yang, Kaizhi Zheng, Ching-Chen Kuo, Yuting Zheng, Sravana~Jyothi Narayanaraju, Xinze Guan, and Xin~Eric Wang.
\newblock Grit: Teaching mllms to think with images.
\newblock \emph{arXiv preprint arXiv:2505.15879}, 2025.

\bibitem[Feng et~al.(2025)Feng, Gong, Li, Guo, Wang, Peng, Wu, Zhang, Wang, and Yue]{feng2025video}
Kaituo Feng, Kaixiong Gong, Bohao Li, Zonghao Guo, Yibing Wang, Tianshuo Peng, Junfei Wu, Xiaoying Zhang, Benyou Wang, and Xiangyu Yue.
\newblock Video-r1: Reinforcing video reasoning in mllms.
\newblock \emph{arXiv preprint arXiv:2503.21776}, 2025.

\bibitem[Guo et~al.(2025)Guo, Yang, Zhang, Song, Zhang, Xu, Zhu, Ma, Wang, Bi, et~al.]{guo2025deepseek}
Daya Guo, Dejian Yang, Haowei Zhang, Junxiao Song, Ruoyu Zhang, Runxin Xu, Qihao Zhu, Shirong Ma, Peiyi Wang, Xiao Bi, et~al.
\newblock Deepseek-r1: Incentivizing reasoning capability in llms via reinforcement learning.
\newblock \emph{arXiv preprint arXiv:2501.12948}, 2025.

\bibitem[Hong et~al.(2023)Hong, Zhen, Chen, Zheng, Du, Chen, and Gan]{hong20233d}
Yining Hong, Haoyu Zhen, Peihao Chen, Shuhong Zheng, Yilun Du, Zhenfang Chen, and Chuang Gan.
\newblock 3d-llm: Injecting the 3d world into large language models.
\newblock \emph{Advances in Neural Information Processing Systems}, 36:\penalty0 20482--20494, 2023.

\bibitem[Huang et~al.(2025)Huang, Jia, Zhai, Cao, Ye, Zhao, Xu, Hu, and Lin]{huang2025vision-r1}
Wenxuan Huang, Bohan Jia, Zijie Zhai, Shaosheng Cao, Zheyu Ye, Fei Zhao, Zhe Xu, Yao Hu, and Shaohui Lin.
\newblock Vision-r1: Incentivizing reasoning capability in multimodal large language models.
\newblock \emph{arXiv preprint arXiv:2503.06749}, 2025.

\bibitem[Hurst et~al.(2024)Hurst, Lerer, Goucher, Perelman, Ramesh, Clark, Ostrow, Welihinda, Hayes, Radford, et~al.]{hurst2024gpt}
Aaron Hurst, Adam Lerer, Adam~P Goucher, Adam Perelman, Aditya Ramesh, Aidan Clark, AJ~Ostrow, Akila Welihinda, Alan Hayes, Alec Radford, et~al.
\newblock Gpt-4o system card.
\newblock \emph{arXiv preprint arXiv:2410.21276}, 2024.

\bibitem[Jiang et~al.(2024)Jiang, Sablayrolles, Roux, Mensch, Savary, Bamford, Chaplot, Casas, Hanna, Bressand, et~al.]{jiang2024mixtral}
Albert~Q Jiang, Alexandre Sablayrolles, Antoine Roux, Arthur Mensch, Blanche Savary, Chris Bamford, Devendra~Singh Chaplot, Diego de~las Casas, Emma~Bou Hanna, Florian Bressand, et~al.
\newblock Mixtral of experts.
\newblock \emph{arXiv preprint arXiv:2401.04088}, 2024.

\bibitem[Kamath et~al.(2023)Kamath, Hessel, and Chang]{kamath2023s}
Amita Kamath, Jack Hessel, and Kai-Wei Chang.
\newblock What's" up" with vision-language models? investigating their struggle with spatial reasoning.
\newblock \emph{arXiv preprint arXiv:2310.19785}, 2023.

\bibitem[Kuhn et~al.(2023)Kuhn, Gal, and Farquhar]{kuhn2023semantic}
Lorenz Kuhn, Yarin Gal, and Sebastian Farquhar.
\newblock Semantic uncertainty: Linguistic invariances for uncertainty estimation in natural language generation.
\newblock \emph{arXiv preprint arXiv:2302.09664}, 2023.

\bibitem[Kullback(1951)]{kullback1951kl}
Solomon Kullback.
\newblock Kullback-leibler divergence.
\newblock \emph{Tech. Rep.}, 1951.

\bibitem[Li et~al.(2024{\natexlab{a}})Li, Zhang, Guo, Zhang, Li, Zhang, Zhang, Zhang, Li, Liu, et~al.]{li2024llava-one}
Bo~Li, Yuanhan Zhang, Dong Guo, Renrui Zhang, Feng Li, Hao Zhang, Kaichen Zhang, Peiyuan Zhang, Yanwei Li, Ziwei Liu, et~al.
\newblock Llava-onevision: Easy visual task transfer.
\newblock \emph{arXiv preprint arXiv:2408.03326}, 2024{\natexlab{a}}.

\bibitem[Li et~al.(2025{\natexlab{a}})Li, Li, Wang, Yan, Zhang, Chen, Hou, Jiang, Zhang, Shen, et~al.]{li2025viewspatial}
Dingming Li, Hongxing Li, Zixuan Wang, Yuchen Yan, Hang Zhang, Siqi Chen, Guiyang Hou, Shengpei Jiang, Wenqi Zhang, Yongliang Shen, et~al.
\newblock Viewspatial-bench: Evaluating multi-perspective spatial localization in vision-language models.
\newblock \emph{arXiv preprint arXiv:2505.21500}, 2025{\natexlab{a}}.

\bibitem[Li et~al.(2024{\natexlab{b}})Li, Zhang, Zhang, Zhang, Li, Li, Ma, and Li]{li2024llava-nxt}
Feng Li, Renrui Zhang, Hao Zhang, Yuanhan Zhang, Bo~Li, Wei Li, Zejun Ma, and Chunyuan Li.
\newblock Llava-next-interleave: Tackling multi-image, video, and 3d in large multimodal models.
\newblock \emph{arXiv preprint arXiv:2407.07895}, 2024{\natexlab{b}}.

\bibitem[Li et~al.(2025{\natexlab{b}})Li, Yan, Meng, Dong, Zeng, He, Wang, Qiao, Wang, and Wang]{li2025videochat}
Xinhao Li, Ziang Yan, Desen Meng, Lu~Dong, Xiangyu Zeng, Yinan He, Yali Wang, Yu~Qiao, Yi~Wang, and Limin Wang.
\newblock Videochat-r1: Enhancing spatio-temporal perception via reinforcement fine-tuning.
\newblock \emph{arXiv preprint arXiv:2504.06958}, 2025{\natexlab{b}}.

\bibitem[Li et~al.(2025{\natexlab{c}})Li, Zhang, Lin, Liu, Cai, Liu, and Zhao]{li2025sti}
Yun Li, Yiming Zhang, Tao Lin, XiangRui Liu, Wenxiao Cai, Zheng Liu, and Bo~Zhao.
\newblock Sti-bench: Are mllms ready for precise spatial-temporal world understanding?
\newblock \emph{arXiv preprint arXiv:2503.23765}, 2025{\natexlab{c}}.

\bibitem[Li et~al.(2025{\natexlab{d}})Li, Yu, Huang, Liu, Liang, Liu, Che, Yu, Boyd-Graber, Mi, et~al.]{li2025self-rewarding}
Zongxia Li, Wenhao Yu, Chengsong Huang, Rui Liu, Zhenwen Liang, Fuxiao Liu, Jingxi Che, Dian Yu, Jordan Boyd-Graber, Haitao Mi, et~al.
\newblock Self-rewarding vision-language model via reasoning decomposition.
\newblock \emph{arXiv preprint arXiv:2508.19652}, 2025{\natexlab{d}}.

\bibitem[Liao et~al.(2025)Liao, Xie, Zhang, Kong, Lu, Yang, and Deng]{liao2025improved}
Zhenyi Liao, Qingsong Xie, Yanhao Zhang, Zijian Kong, Haonan Lu, Zhenyu Yang, and Zhijie Deng.
\newblock Improved visual-spatial reasoning via r1-zero-like training.
\newblock \emph{arXiv preprint arXiv:2504.00883}, 2025.

\bibitem[Lin et~al.(2014)Lin, Maire, Belongie, Hays, Perona, Ramanan, Doll{\'a}r, and Zitnick]{lin2014coco}
Tsung-Yi Lin, Michael Maire, Serge Belongie, James Hays, Pietro Perona, Deva Ramanan, Piotr Doll{\'a}r, and C~Lawrence Zitnick.
\newblock Microsoft coco: Common objects in context.
\newblock In \emph{European conference on computer vision}, pp.\  740--755. Springer, 2014.

\bibitem[Liu et~al.(2025{\natexlab{a}})Liu, Dong, Wang, Ma, Tang, Tang, Rao, Ma, and Krishna]{liu2025coarse}
Benlin Liu, Yuhao Dong, Yiqin Wang, Zixian Ma, Yansong Tang, Luming Tang, Yongming Rao, Wei-Chiu Ma, and Ranjay Krishna.
\newblock Coarse correspondences boost spatial-temporal reasoning in multimodal language model.
\newblock In \emph{Proceedings of the Computer Vision and Pattern Recognition Conference}, pp.\  3783--3792, 2025{\natexlab{a}}.

\bibitem[Liu et~al.(2025{\natexlab{b}})Liu, Sun, Zang, Dong, Cao, Duan, Lin, and Wang]{liu2025visual-rft}
Ziyu Liu, Zeyi Sun, Yuhang Zang, Xiaoyi Dong, Yuhang Cao, Haodong Duan, Dahua Lin, and Jiaqi Wang.
\newblock Visual-rft: Visual reinforcement fine-tuning.
\newblock \emph{arXiv preprint arXiv:2503.01785}, 2025{\natexlab{b}}.

\bibitem[Meng et~al.(2025)Meng, Du, Liu, Zhou, Lu, Fu, Shi, Wang, He, Zhang, et~al.]{meng2025mm}
Fanqing Meng, Lingxiao Du, Zongkai Liu, Zhixiang Zhou, Quanfeng Lu, Daocheng Fu, Botian Shi, Wenhai Wang, Junjun He, Kaipeng Zhang, et~al.
\newblock Mm-eureka: Exploring visual aha moment with rule-based large-scale reinforcement learning.
\newblock \emph{CoRR}, 2025.

\bibitem[Ouyang et~al.(2025)Ouyang, Liu, Wu, Liu, Zhou, Zhou, Meng, and Sun]{ouyang2025spacer}
Kun Ouyang, Yuanxin Liu, Haoning Wu, Yi~Liu, Hao Zhou, Jie Zhou, Fandong Meng, and Xu~Sun.
\newblock Spacer: Reinforcing mllms in video spatial reasoning.
\newblock \emph{arXiv preprint arXiv:2504.01805}, 2025.

\bibitem[Ouyang et~al.(2022)Ouyang, Wu, Jiang, Almeida, Wainwright, Mishkin, Zhang, Agarwal, Slama, Ray, et~al.]{ouyang2022sft}
Long Ouyang, Jeffrey Wu, Xu~Jiang, Diogo Almeida, Carroll Wainwright, Pamela Mishkin, Chong Zhang, Sandhini Agarwal, Katarina Slama, Alex Ray, et~al.
\newblock Training language models to follow instructions with human feedback.
\newblock \emph{Advances in neural information processing systems}, 35:\penalty0 27730--27744, 2022.

\bibitem[Shao et~al.(2024)Shao, Wang, Zhu, Xu, Song, Bi, Zhang, Zhang, Li, Wu, et~al.]{shao2024deepseekmath}
Zhihong Shao, Peiyi Wang, Qihao Zhu, Runxin Xu, Junxiao Song, Xiao Bi, Haowei Zhang, Mingchuan Zhang, YK~Li, Yang Wu, et~al.
\newblock Deepseekmath: Pushing the limits of mathematical reasoning in open language models.
\newblock \emph{arXiv preprint arXiv:2402.03300}, 2024.

\bibitem[Shen et~al.(2025)Shen, Liu, Li, Fang, Ma, Liao, Shen, Zhang, Zhao, Zhang, et~al.]{shen2025vlm-r1}
Haozhan Shen, Peng Liu, Jingcheng Li, Chunxin Fang, Yibo Ma, Jiajia Liao, Qiaoli Shen, Zilun Zhang, Kangjia Zhao, Qianqian Zhang, et~al.
\newblock Vlm-r1: A stable and generalizable r1-style large vision-language model.
\newblock \emph{arXiv preprint arXiv:2504.07615}, 2025.

\bibitem[Su et~al.(2025{\natexlab{a}})Su, Wang, Ren, Lin, and Chen]{su2025pixel}
Alex Su, Haozhe Wang, Weiming Ren, Fangzhen Lin, and Wenhu Chen.
\newblock Pixel reasoner: Incentivizing pixel-space reasoning with curiosity-driven reinforcement learning.
\newblock \emph{arXiv preprint arXiv:2505.15966}, 2025{\natexlab{a}}.

\bibitem[Su et~al.(2025{\natexlab{b}})Su, Xia, Guo, Liu, Ma, Qu, Liu, Li, Zeng, Yang, et~al.]{su2025thinking_with_image}
Zhaochen Su, Peng Xia, Hangyu Guo, Zhenhua Liu, Yan Ma, Xiaoye Qu, Jiaqi Liu, Yanshu Li, Kaide Zeng, Zhengyuan Yang, et~al.
\newblock Thinking with images for multimodal reasoning: Foundations, methods, and future frontiers.
\newblock \emph{arXiv preprint arXiv:2506.23918}, 2025{\natexlab{b}}.

\bibitem[Team et~al.(2024)Team, Georgiev, Lei, Burnell, Bai, Gulati, Tanzer, Vincent, Pan, Wang, et~al.]{team2024gemini}
Gemini Team, Petko Georgiev, Ving~Ian Lei, Ryan Burnell, Libin Bai, Anmol Gulati, Garrett Tanzer, Damien Vincent, Zhufeng Pan, Shibo Wang, et~al.
\newblock Gemini 1.5: Unlocking multimodal understanding across millions of tokens of context.
\newblock \emph{arXiv preprint arXiv:2403.05530}, 2024.

\bibitem[Team et~al.(2025)Team, Du, Yin, Xing, Qu, Wang, Chen, Zhang, Du, Wei, et~al.]{team2025kimi}
Kimi Team, Angang Du, Bohong Yin, Bowei Xing, Bowen Qu, Bowen Wang, Cheng Chen, Chenlin Zhang, Chenzhuang Du, Chu Wei, et~al.
\newblock Kimi-vl technical report.
\newblock \emph{arXiv preprint arXiv:2504.07491}, 2025.

\bibitem[Tian et~al.(2024)Tian, Gu, Li, Liu, Wang, Zhao, Zhan, Jia, Lang, and Zhao]{tian2024drivevlm}
Xiaoyu Tian, Junru Gu, Bailin Li, Yicheng Liu, Yang Wang, Zhiyong Zhao, Kun Zhan, Peng Jia, Xianpeng Lang, and Hang Zhao.
\newblock Drivevlm: The convergence of autonomous driving and large vision-language models.
\newblock \emph{arXiv preprint arXiv:2402.12289}, 2024.

\bibitem[Tong et~al.(2024)Tong, Brown, Wu, Woo, IYER, Akula, Yang, Yang, Middepogu, Wang, et~al.]{tong2024cambrian}
Peter Tong, Ellis Brown, Penghao Wu, Sanghyun Woo, Adithya Jairam~Vedagiri IYER, Sai~Charitha Akula, Shusheng Yang, Jihan Yang, Manoj Middepogu, Ziteng Wang, et~al.
\newblock Cambrian-1: A fully open, vision-centric exploration of multimodal llms.
\newblock \emph{Advances in Neural Information Processing Systems}, 37:\penalty0 87310--87356, 2024.

\bibitem[Wang et~al.(2025)Wang, Chen, Karaev, Vedaldi, Rupprecht, and Novotny]{wang2025vggt}
Jianyuan Wang, Minghao Chen, Nikita Karaev, Andrea Vedaldi, Christian Rupprecht, and David Novotny.
\newblock Vggt: Visual geometry grounded transformer.
\newblock In \emph{Proceedings of the Computer Vision and Pattern Recognition Conference}, pp.\  5294--5306, 2025.

\bibitem[Wei et~al.(2022)Wei, Wang, Schuurmans, Bosma, Xia, Chi, Le, Zhou, et~al.]{wei2022cot}
Jason Wei, Xuezhi Wang, Dale Schuurmans, Maarten Bosma, Fei Xia, Ed~Chi, Quoc~V Le, Denny Zhou, et~al.
\newblock Chain-of-thought prompting elicits reasoning in large language models.
\newblock \emph{Advances in neural information processing systems}, 35:\penalty0 24824--24837, 2022.

\bibitem[Wu et~al.(2025{\natexlab{a}})Wu, Liu, Hung, and Duan]{wu2025spatial}
Diankun Wu, Fangfu Liu, Yi-Hsin Hung, and Yueqi Duan.
\newblock Spatial-mllm: Boosting mllm capabilities in visual-based spatial intelligence.
\newblock \emph{arXiv preprint arXiv:2505.23747}, 2025{\natexlab{a}}.

\bibitem[Wu et~al.(2025{\natexlab{b}})Wu, Huang, Chen, Zhang, Wang, and Xie]{wu2025spatialscore}
Haoning Wu, Xiao Huang, Yaohui Chen, Ya~Zhang, Yanfeng Wang, and Weidi Xie.
\newblock Spatialscore: Towards unified evaluation for multimodal spatial understanding.
\newblock \emph{arXiv preprint arXiv:2505.17012}, 2025{\natexlab{b}}.

\bibitem[Wu et~al.(2025{\natexlab{c}})Wu, Guan, Feng, Liu, Wu, Wang, Wu, and Tan]{wu2025reinforcing}
Junfei Wu, Jian Guan, Kaituo Feng, Qiang Liu, Shu Wu, Liang Wang, Wei Wu, and Tieniu Tan.
\newblock Reinforcing spatial reasoning in vision-language models with interwoven thinking and visual drawing.
\newblock \emph{arXiv preprint arXiv:2506.09965}, 2025{\natexlab{c}}.

\bibitem[Yang et~al.(2025{\natexlab{a}})Yang, Yang, Gupta, Han, Fei-Fei, and Xie]{yang2025thinking}
Jihan Yang, Shusheng Yang, Anjali~W Gupta, Rilyn Han, Li~Fei-Fei, and Saining Xie.
\newblock Thinking in space: How multimodal large language models see, remember, and recall spaces.
\newblock In \emph{Proceedings of the Computer Vision and Pattern Recognition Conference}, pp.\  10632--10643, 2025{\natexlab{a}}.

\bibitem[Yang et~al.(2025{\natexlab{b}})Yang, He, Pan, Jiang, Deng, Yang, Lu, Yin, Rao, Zhu, et~al.]{yang2025r1-onevision}
Yi~Yang, Xiaoxuan He, Hongkun Pan, Xiyan Jiang, Yan Deng, Xingtao Yang, Haoyu Lu, Dacheng Yin, Fengyun Rao, Minfeng Zhu, et~al.
\newblock R1-onevision: Advancing generalized multimodal reasoning through cross-modal formalization.
\newblock \emph{arXiv preprint arXiv:2503.10615}, 2025{\natexlab{b}}.

\bibitem[Yeshwanth et~al.(2023)Yeshwanth, Liu, Nie{\ss}ner, and Dai]{yeshwanth2023scannet++}
Chandan Yeshwanth, Yueh-Cheng Liu, Matthias Nie{\ss}ner, and Angela Dai.
\newblock Scannet++: A high-fidelity dataset of 3d indoor scenes.
\newblock In \emph{Proceedings of the IEEE/CVF International Conference on Computer Vision}, pp.\  12--22, 2023.

\bibitem[Yu et~al.(2025)Yu, Lin, Zhao, Yin, Wei, Peng, Wei, Sun, Han, Ge, et~al.]{yu2025perception}
En~Yu, Kangheng Lin, Liang Zhao, Jisheng Yin, Yana Wei, Yuang Peng, Haoran Wei, Jianjian Sun, Chunrui Han, Zheng Ge, et~al.
\newblock Perception-r1: Pioneering perception policy with reinforcement learning.
\newblock \emph{arXiv preprint arXiv:2504.07954}, 2025.

\bibitem[Zhang et~al.(2025)Zhang, Chen, Zhou, Xu, Huang, Mei, Chen, Yuan, Cai, Huang, et~al.]{zhang2025flatland}
Jiahui Zhang, Yurui Chen, Yanpeng Zhou, Yueming Xu, Ze~Huang, Jilin Mei, Junhui Chen, Yu-Jie Yuan, Xinyue Cai, Guowei Huang, et~al.
\newblock From flatland to space: Teaching vision-language models to perceive and reason in 3d.
\newblock \emph{arXiv preprint arXiv:2503.22976}, 2025.

\bibitem[Zheng et~al.(2025)Zheng, Huang, and Wang]{zheng2025video-3d}
Duo Zheng, Shijia Huang, and Liwei Wang.
\newblock Video-3d llm: Learning position-aware video representation for 3d scene understanding.
\newblock In \emph{Proceedings of the Computer Vision and Pattern Recognition Conference}, pp.\  8995--9006, 2025.

\bibitem[Zhou et~al.(2017)Zhou, Zhao, Puig, Fidler, Barriuso, and Torralba]{zhou2017ade20k}
Bolei Zhou, Hang Zhao, Xavier Puig, Sanja Fidler, Adela Barriuso, and Antonio Torralba.
\newblock Scene parsing through ade20k dataset.
\newblock In \emph{Proceedings of the IEEE conference on computer vision and pattern recognition}, pp.\  633--641, 2017.

\bibitem[Zhu et~al.(2024)Zhu, Wang, Zhang, Pang, and Liu]{zhu2024llava-3d}
Chenming Zhu, Tai Wang, Wenwei Zhang, Jiangmiao Pang, and Xihui Liu.
\newblock Llava-3d: A simple yet effective pathway to empowering lmms with 3d-awareness.
\newblock \emph{arXiv preprint arXiv:2409.18125}, 2024.

\bibitem[Zitkovich et~al.(2023)Zitkovich, Yu, Xu, Xu, Xiao, Xia, Wu, Wohlhart, Welker, Wahid, et~al.]{zitkovich2023rt}
Brianna Zitkovich, Tianhe Yu, Sichun Xu, Peng Xu, Ted Xiao, Fei Xia, Jialin Wu, Paul Wohlhart, Stefan Welker, Ayzaan Wahid, et~al.
\newblock Rt-2: Vision-language-action models transfer web knowledge to robotic control.
\newblock In \emph{Conference on Robot Learning}, pp.\  2165--2183. PMLR, 2023.

\end{thebibliography}
